\documentclass[sn-mathphys,Numbered]{sn-jnl}

\usepackage{graphicx}%
\usepackage{multirow}%
\usepackage{amsmath,amssymb,amsfonts}%
\usepackage{amsthm}%
\usepackage{mathrsfs}%
\usepackage[title]{appendix}%
\usepackage{xcolor}%
\usepackage{textcomp}%
\usepackage{manyfoot}%
\usepackage{booktabs}%
\usepackage{algorithm}%
\usepackage{algorithmicx}%
\usepackage{algpseudocode}%
\usepackage{listings}%
\usepackage{graphicx}
\usepackage{floatflt}
\usepackage{tikz}

\newtheorem{theorem}{Theorem}
\newtheorem{proposition}[theorem]{Proposition}%
\newtheorem{example}{Example}%
\newtheorem{definition}{Definition}%

\DeclareMathOperator*{\argmin}{argmin} 

\begin{document}

\title[Quantifying Harm]{Quantifying Harm\footnote{Preprint: currently under review}}



\author[1]{\fnm{Sander} \sur{Beckers}}\email{srekcebrednas@gmail.com}

\author[2]{\fnm{Hana} \sur{Chockler}}\email{hana.chockler@kcl.ac.uk}

\author[3]{\fnm{Joseph Y.} \sur{Halpern}}
\equalcont{Deceased: 02/13/2026}

\affil[1]{\orgdiv{Department of Statistical Science}, \orgname{University College London}
}

\affil[2]{\orgdiv{Department of Informatics}, \orgname{King's College London} 
}

\affil[3]{\orgdiv{Department of Computer Science}, \orgname{Cornell University}
}



\abstract{
  In earlier work
we defined  a qualitative notion of harm:
either harm is caused, or it is not. 
For practical applications, we often need to quantify harm; for
example, we may want to choose the least harmful  of a set of possible
interventions.
In this work, which is an expanded version of the conference paper
\cite{BCH22b}, we develop a quantitative notion of harm. 
We first present a quantitative definition of harm in a deterministic
context involving a single individual, then we consider the issues involved in dealing with uncertainty
regarding the context and going from a notion of harm for a single
individual to a notion of ``societal harm'', which involves
aggregating the harm to individuals.  We show that the ``obvious'' way
of doing this (just taking the expected harm for an individual and
then summing the expected harm over all individuals) can lead to
counterintuitive or inappropriate answers, and discuss alternatives,
drawing on work from the decision-theory literature.
Finally, we connect our work to a recent debate over harm within the context of precision medicine. 
}




\maketitle


\newcommand{\Pro}{\textit{Pr}}
\newcommand{\penn}{\textit{penn}}
\newcommand{\texas}{\textit{texas}}
\newcommand{\idle}{\textit{idle}}
\newcommand{\funcv}{\textit{F}}
\newcommand{\sign}{{\cal S}}
\newcommand{\func}{{\cal F}}
\newcommand{\rel}{{\cal R}}
\newcommand{\msign}{${\cal S}$ }
\newcommand{\mfunc}{${\cal F}$ }
\newcommand{\mmod}{${\cal M}$ }
\newcommand{\exog}{{\cal U}}
\newcommand{\mexog}{${\cal U}$ }
\newcommand{\sexog}{ \vec{u}}
\newcommand{\smexog}{$\vec{u}$ }
\newcommand{\sexogtwo}{ \vec{c}}
\newcommand{\smexogtwo}{$\vec{c}$ }
\newcommand{\mvar}{${\cal V}$ }
\newcommand{\var}{{\cal V}}
\newcommand{\sm}{\textit{sim}}
\newcommand{\comp}{\textit{comp}}

\newcommand{\arson}{{\cal A}}
\newcommand{\malfunc}{{\cal EM}}
\newcommand{\ws}{{\cal WS}}
\newcommand{\resdeath}{\textit{RD}}
\newcommand{\cs}{{\cal CS}}
\newcommand{\Scal}{{\cal S}}

\newcommand{\NAr}{\mbox{{\it NA}}\xspace}
\newcommand{\OV}{\mbox{{\it OV}}\xspace}
\newcommand{\SV}{\mbox{{\it SV}}\xspace}
\newcommand{\GS}{\mbox{{\it GS}}\xspace}
\newcommand{\GL}{\mbox{{\it GL}}\xspace}
\newcommand{\SF}{\mbox{{\it S}}\xspace}
\newcommand{\FA}{\mbox{{\it FA}}\xspace}
\newcommand{\FIRE}{\mbox{{\it FI}}\xspace}
\newcommand{\SO}{\mbox{{\it SO}}\xspace}
\newcommand{\RD}{\mbox{{\it RD}}\xspace}
\newcommand{\scRD}{\mbox{{\scriptsize{\it RD}}}}
\newcommand{\CS}{\mbox{{\it CS}}\xspace}
\newcommand{\WS}{\mbox{{\it WS}}\xspace}
\newcommand{\E}{\mbox{{\it E}}\xspace}
\newcommand{\scWS}{\mbox{{\scriptsize{\it WS}}}}
\newcommand{\scVEN}{\mbox{{\scriptsize{\it V}}}}
\newcommand{\SM}{\mbox{{\it SM}}}
\newcommand{\VEN}{\mbox{{\it V}}\xspace}

\newcommand{\SH}{\mbox{{\it SH}}}
\newcommand{\BH}{\mbox{{\it BH}}}
\newcommand{\BT}{\mbox{{\it BT}}}
\newcommand{\BS}{\mbox{{\it BS}}}
\newcommand{\ST}{\mbox{{\it ST}}}
\newcommand{\EX}{\mbox{{\it EX}}}
\newcommand{\EF}{\mbox{{\it EF}}}
\newcommand{\AG}{\mbox{{\it AG}}}
\newcommand{\AF}{\mbox{{\it AF}}}

\newcommand{\VG}{\mbox{{\it VG}}}
\newcommand{\VB}{\mbox{{\it VB}}}
\newcommand{\OP}{\mbox{{\it OP}}}
\newcommand{\AT}{\mbox{{\it AT}}}

\newcommand{\CA}{\mbox{{\it C}}}
\newcommand{\DV}{\mbox{{\it DV}}}
\newcommand{\FS}{\mbox{{\it FS}}}
\newcommand{\IS}{\mbox{{\it IS}}}
\newcommand{\RV}{\mbox{{\it RV}}}
\newcommand{\RA}{\mbox{{\it A}}}
\newcommand{\DT}{\mbox{{\it D}}}
\newcommand{\AP}{\mbox{{\it AP}}}
\newcommand{\PLS}{\mbox{{\it PLS}}}

\newcommand{\sPD}{\mbox{{\scriptsize{\it PD}}}}
\newcommand{\sRC}{\mbox{{\scriptsize{{\it R}}}}}
\newcommand{\sIL}{\mbox{{\scriptsize{{\it IL}}}}}
\newcommand{\sSG}{\mbox{{\scriptsize{{\it SG}}}}}
\newcommand{\sTT}{\mbox{{\scriptsize{{\it T}}}}}
\newcommand{\sDP}{\mbox{{\scriptsize{{\it D}}}}}
\newcommand{\sDO}{\mbox{{\scriptsize{{\it PD}}}}}
\newcommand{\sAS}{\mbox{{\scriptsize{{\it AS}}}}}
\newcommand{\sRS}{\mbox{{\scriptsize{{\it RS}}}}}
\newcommand{\sDE}{\mbox{{\scriptsize{{\it D}}}}}
\newcommand{\sMC}{\mbox{{\scriptsize{{\it C}}}}}
\newcommand{\sUW}{\mbox{{\scriptsize{{\it W}}}}}
\newcommand{\sFM}{\mbox{{\scriptsize{{\it FM}}}}}
\newcommand{\sLM}{\mbox{{\scriptsize{{\it LM}}}}}
\newcommand{\sGC}{\mbox{{\scriptsize{{\it CG}}}}}
\newcommand{\sCM}{\mbox{{\scriptsize{{\it CM}}}}}
\newcommand{\sAM}{\mbox{{\scriptsize{{\it AM}}}}}
\newcommand{\sSC}{\mbox{{\scriptsize{{\it SC}}}}}
\newcommand{\sRG}{\mbox{{\scriptsize{{\it RG}}}}}
\newcommand{\sCB}{\mbox{{\scriptsize{{\it CB}}}}}

\newcommand{\PD}{\mbox{{\it PD}}}
\newcommand{\RC}{\mbox{{\it R}}}
\newcommand{\IL}{\mbox{{\it IL}}}
\newcommand{\MO}{\mbox{{\it MO}}}
\newcommand{\SG}{\mbox{{\it SG}}}
\newcommand{\IT}{\mbox{{\it IT}}}
\newcommand{\IM}{\mbox{{\it IM}}}
\newcommand{\TT}{\mbox{{\it T}}}
\newcommand{\DP}{\mbox{{\it D}}}
\newcommand{\MT}{\mbox{{\it MT}}}
\newcommand{\ER}{\mbox{{\it ER}}}
\newcommand{\EP}{\mbox{{\it EP}}}
\newcommand{\DO}{\mbox{{\it PD}}}
\newcommand{\AS}{\mbox{{\it AS}}}
\newcommand{\NS}{\mbox{{\it NS}}}
\newcommand{\RS}{\mbox{{\it RS}}}
\newcommand{\DE}{\mbox{{\it D}}}
\newcommand{\MC}{\mbox{{\it C}}}
\newcommand{\UW}{\mbox{{\it W}}}
\newcommand{\FM}{\mbox{{\it FM}}}
\newcommand{\LM}{\mbox{{\it LM}}}
\newcommand{\GC}{\mbox{{\it CG}}}
\newcommand{\LC}{\mbox{{\it LC}}}
\newcommand{\IG}{\mbox{{\it IG}}}
\newcommand{\HL}{\mbox{{\it HL}}}
\newcommand{\Con}{\mbox{{\it C}}}
\newcommand{\NR}{\mbox{{\it NR}}}
\newcommand{\MA}{\mbox{{\it MA}}}
\newcommand{\CM}{\mbox{{\it CM}}}
\newcommand{\AM}{\mbox{{\it AM}}}
\newcommand{\SC}{\mbox{{\it SC}}}
\newcommand{\RG}{\mbox{{\it RG}}}
\newcommand{\CB}{\mbox{{\it CB}}}

\newcommand{\subq}{q \rightarrow {\neg{q}}}
\newcommand{\ioi}{if and only if }
\newcommand{\es}{\emptyset}
\newcommand{\bd}{\begin{definition}}
\newcommand{\ed}{\end{definition}}
\newcommand{\be}{\begin{enumerate}}
\newcommand{\bi}{\begin{itemize}}
\newcommand{\ee}{\end{enumerate}}
\newcommand{\ei}{\end{itemize}}

\newcommand{\N}{\mbox{I$\!$N}}
\newcommand{\A}{{\cal A}}
\newcommand{\U}{{\cal U}}
\newcommand{\D}{{D}}
\newcommand{\I}{{\cal I}}
\newcommand{\cS}{{\cal S}}
\newcommand{\C}{{\cal C}}
\newcommand{\Lan}{L}
\newcommand{\G}{{\cal G}}
\newcommand{\cF}{{\cal F}}
\newcommand{\V}{{\cal V}}
\newcommand{\R}{{\cal R}}
\newcommand{\Z}{{\cal Z}}
\newcommand{\W}{{\cal W}}
\newcommand{\Wt}{{\mathcal W}\textit{eight}}
\newcommand{\zug}[1]{\langle #1  \rangle}
\newcommand{\twin}{\it twin}
\newcommand{\stam}[1]{}
\newcommand{\bft}{{\bf true}\xspace}   
\newcommand{\bff}{{\bf false}\xspace}
\newcommand{\Lfair}{L_{\mbox{\it fair}}}

\newcommand{\commentout}[1]{}

\newcommand{\fullv}[1]{#1}
\newcommand{\shortv}[1]{\commentout{#1}}
\newcommand{\abst}[1]{#1}

\newcommand{\dr}{\mbox{{\em dr}}}
\newcommand{\db}{\mbox{{\em db}}}
\newcommand{\K}{{\cal K}}
\newcommand{\kset}{{\cal k}}
\newcommand{\NP}{\mbox{{\it NP}}}

\newtheorem{property}{Property}
\newtheorem{lemma}{Lemma}
\newtheorem{observation}{Observation}
\newtheorem{claim}{Claim}
\newcommand{\prf}{\noindent{\bf Proof.} }

\newcommand{\wbox}{\mbox{$\sqcap$\llap{$\sqcup$}}}
\newcommand{\bbox}{\vrule height7pt width4pt depth1pt}

\newcommand{\clm}{\begin{claim}}
\newcommand{\eclm}{\end{claim}}
\let\member=\in
\let\notmember=\notin
\newcommand{\sub}{_}
\def\su{^}
\newcommand{\rarrow}{\rightarrow}
\newcommand{\larrow}{\leftarrow}
\newcommand{\bolda}{{\bf a}}
\newcommand{\boldb}{{\bf b}}
\newcommand{\boldc}{{\bf c}}
\newcommand{\boldd}{{\bf d}}
\newcommand{\bolde}{{\bf e}}
\newcommand{\boldf}{{\bf f}}
\newcommand{\boldg}{{\bf g}}
\newcommand{\boldh}{{\bf h}}
\newcommand{\boldi}{{\bf i}}
\newcommand{\boldj}{{\bf j}}
\newcommand{\boldk}{{\bf k}}
\newcommand{\boldl}{{\bf l}}
\newcommand{\boldm}{{\bf m}}
\newcommand{\boldn}{{\bf n}}
\newcommand{\boldo}{{\bf o}}
\newcommand{\boldp}{{\bf p}}
\newcommand{\boldq}{{\bf q}}
\newcommand{\boldr}{{\bf r}}
\newcommand{\bolds}{{\bf s}}
\newcommand{\boldt}{{\bf t}}
\newcommand{\boldu}{{\bf u}}
\newcommand{\boldv}{{\bf v}}
\newcommand{\boldw}{{\bf w}}
\newcommand{\boldx}{{\bf x}}
\newcommand{\boldy}{{\bf y}}
\newcommand{\boldz}{{\bf z}}
\newcommand{\boldA}{{\bf A}}
\newcommand{\boldB}{{\bf B}}
\newcommand{\boldC}{{\bf C}}
\newcommand{\boldD}{{\bf D}}
\newcommand{\boldE}{{\bf E}}
\newcommand{\boldF}{{\bf F}}
\newcommand{\boldG}{{\bf G}}
\newcommand{\boldH}{{\bf H}}
\newcommand{\boldI}{{\bf I}}
\newcommand{\boldJ}{{\bf J}}
\newcommand{\boldK}{{\bf K}}
\newcommand{\boldL}{{\bf L}}
\newcommand{\boldM}{{\bf M}}
\newcommand{\boldN}{{\bf N}}
\newcommand{\boldO}{{\bf O}}
\newcommand{\boldP}{{\bf P}}
\newcommand{\boldQ}{{\bf Q}}
\newcommand{\boldR}{{\bf R}}
\newcommand{\boldS}{{\bf S}}
\newcommand{\boldT}{{\bf T}}
\newcommand{\boldU}{{\bf U}}
\newcommand{\boldV}{{\bf V}}
\newcommand{\boldW}{{\bf W}}
\newcommand{\boldX}{{\bf X}}
\newcommand{\boldY}{{\bf Y}}
\newcommand{\boldZ}{{\bf Z}}

\newcommand{\dtur}{\modls}
\newcommand{\infers}{\vdash}
\newcommand{\stur}{\vdash}
\newcommand{\rimp}{\Rightarrow}
\newcommand{\limp}{\Leftarrow}
\newcommand{\dimp}{\Leftrightarrow}
\newcommand{\bor}{\bigvee}
\newcommand{\band}{\bigwedge}
\newcommand{\union}{\cup}
\newcommand{\inter}{\cap}
\newcommand{\xx}{{\bf x}}
\newcommand{\yy}{{\bf y}}
\newcommand{\uu}{{\bf u}}
\newcommand{\vv}{{\bf v}}
\newcommand{\FF}{{\bf F}}
\newcommand{\natnum}{{\sl N}}
\newcommand{\IR}{\mbox{$I\!\!R$}}
\newcommand{\IP}{\mbox{$I\!\!P$}}
\newcommand{\IN}{\mbox{$I\!\!N$}}
\newcommand{\IC}{\mbox{$C\!\!\!\!\raisebox{.75pt}{\mbox{\sqi I}}$}}
\newcommand{\marrow}{\hbox{$\rightarrow$ \hskip -10pt
                      $\rightarrow$ \hskip 3pt}}
\newcommand{\Circ}{\mbox{{\small $\bigcirc$}}}
\newcommand{\lt}{<}
\newcommand{\gt}{>}
\newcommand{\all}{\forall}
\newcommand{\infinity}{\infty}
\newcommand{\bc}[2]{\left( \begin{array}{c} #1 \\ #2 \end{array} \right)}
\newcommand{\cross}{\times}


\newcommand{\imp}{\Rightarrow}


\newcommand{\B}{{\cal B}}
\newcommand{\cC}{{\cal C}}
\newcommand{\J}{{\cal J}}
\newcommand{\M}{{\cal M}}
\newcommand{\mM}{${\mathcal M}$}
\newcommand{\Ocal}{{\cal O}}
\newcommand{\Hcal}{{\cal H}}
\newcommand{\mH}{${\mathcal H}$}
\newcommand{\Pcal}{{\cal P}}
\newcommand{\Q}{{\cal Q}}
\newcommand{\T}{{\cal T}}
\newcommand{\mU}{${\mathcal U}$}
\newcommand{\mV}{${\mathcal V}$}
\newcommand{\X}{{\cal X}}
\newcommand{\Y}{{\cal Y}}

\newcommand{\Kone}{{\cal K}_1}
\newcommand{\abs}[1]{\left| #1\right|}
\newcommand{\set}[1]{\left\{ #1 \right\}}
\newcommand{\Ki}{{\cal K}_i}
\newcommand{\Kn}{{\cal K}_n}
\newcommand{\st}{\, \vert \,} 
\newcommand{\stc}{\, : \,} 
\newcommand{\la}{\langle}
\newcommand{\ra}{\rangle}
\newcommand{\<}{\langle}
\newcommand{\lang}{\mbox{${\cal L}_n$}}
\newcommand{\langd}{\mbox{${\cal L}_n^D$}}

\newtheorem{nlem}{Lemma}
\newtheorem{Ob}{Observation}
\newtheorem{pps}{Proposition}
\newtheorem{defn}{Definition}
\newtheorem{crl}{Corollary}
\newtheorem{cl}{Claim}
\newcommand{\pf}{\par\noindent{\bf Proof}~~}
\newcommand{\eg}{e.g.,~}
\newcommand{\ie}{i.e.,~}
\newcommand{\vs}{vs.~}
\newcommand{\cf}{cf.~}
\newcommand{\etal}{et al.\ }
\newcommand{\resp}{resp.\ }
\newcommand{\respc}{resp.,\ }
\newcommand{\comment}[1]{\marginpar{\scriptsize\raggedright #1}}
\newcommand{\wrt}{with respect to~}
\newcommand{\re}{r.e.}
\newcommand{\nind}{\noindent}
\newcommand{\distributed}{distributed\ }
\newcommand{\bn}{\addcontentsline{toc}{section}{Notes}
\bigskip\markright{NOTES}
\section*{Notes}}
\newcommand{\Exer}{
\bigskip\markright{EXERCISES}
\section*{Exercises}}
\newcommand{\DG}{D_G}
\newcommand{\Sm}{{\rm S5}_m}
\newcommand{\Smc}{{\rm S5C}_m}
\newcommand{\Smi}{{\rm S5I}_m}
\newcommand{\Smic}{{\rm S5CI}_m}
\newcommand{\Martin}{Mart\'\i n\ }
\newtheorem{fthm}{Theorem}
\newtheorem{flem}[fthm]{Lemma}
\newtheorem{fcor}[fthm]{Corollary}
\newcommand{\subG}{_G}

\newcommand{\Kax}{{\rm K}_n}
\newcommand{\Kaxc}{{\rm K}_n^C}
\newcommand{\Kaxd}{{\rm K}_n^D}
\newcommand{\Tax}{{\rm T}_n}
\newcommand{\Taxc}{{\rm T}_n^C}
\newcommand{\Taxd}{{\rm T}_n^D}
\newcommand{\fourax}{{\rm S4}_n}
\newcommand{\fouraxc}{{\rm S4}_n^C}
\newcommand{\fouraxd}{{\rm S4}_n^D}
\newcommand{\fiveax}{{\rm S5}_n}
\newcommand{\fiveaxc}{{\rm S5}_n^C}
\newcommand{\fiveaxd}{{\rm S5}_n^D}
\newcommand{\Dax}{{\rm KD45}_n}
\newcommand{\Daxc}{{\rm KD45}_n^C}
\newcommand{\Daxd}{{\rm KD45}_n^D}
\newcommand{\LP}{{\cal L}_n}
\newcommand{\LCP}{{\cal L}_n^C}
\newcommand{\LDP}{{\cal L}_n^D}
\newcommand{\LCDP}{{\cal L}_n^{CD}}
\newcommand{\MP}{{\cal M}_n}
\newcommand{\MPr}{{\cal M}_n^r}
\newcommand{\MPrt}{\M_n^{\mbox{\scriptsize{{\it rt}}}}}
\newcommand{\MPrst}{\M_n^{\mbox{\scriptsize{{\it rst}}}}}
\newcommand{\MPelt}{\M_n^{\mbox{\scriptsize{{\it elt}}}}}
\newcommand{\fiveaxdu}{{\rm S5}_n^{DU}}
\newcommand{\LPD}{{\cal L}_n^D}
\newcommand{\fiveaxu}{{\rm S5}_n^U}
\newcommand{\fiveaxcu}{{\rm S5}_n^{CU}}
\newcommand{\LPU}{{\cal L}^{U}_n}
\newcommand{\LPCU}{{\cal L}_n^{CU}}
\newcommand{\LDPU}{{\cal L}_n^{DU}}
\newcommand{\LCPU}{{\cal L}_n^{CU}}
\newcommand{\LPDU}{{\cal L}_n^{DU}}
\newcommand{\LPCDU}{{\cal L}_n^{\it CDU}}
\newcommand{\Cn}{\C_n}
\newcommand{\CSnp}{\I_n^{oa}(\Phi')}
\newcommand{\CSc}{\C_n^{oa}(\Phi)}
\newcommand{\Ccs}{\C_n^{oa}}
\newcommand{\CSAX}{OA$_{n,\Phi}$}
\newcommand{\CSAXN}{OA$_{n,{\Phi}}'$}
\newcommand{\untill}{U}
\newcommand{\until}{\, U \,}
\newcommand{\amp}{{\rm a.m.p.}}

\newcommand{\msgc}[1]{ @ #1 }
\newcommand{\Camp}{{\C_n^{\it amp}}}

\newcommand{\thm}{\begin{theorem}}
\newcommand{\pro}{\begin{proposition}}
\newcommand{\ethm}{\end{theorem}}

\newcommand{\fpa}{$\mbox{FP}^{{\rm A}[\log{n}]}$}
\newcommand{\fp}{$\mbox{FP}^{{\rm NP}[\log{n}]}$}
\renewcommand{\phi}{\varphi}

\newcommand{\rarrowr}{\stackrel{r}{\rightarrow}}
\newcommand{\Gz}{\G_0}
\newcommand{\denselist}{\itemsep 0pt\partopsep 0pt}
\def\seealso#1#2{({\em see also\/} #1), #2}
\newcommand{\cents}{\hbox{\rm \rlap{/}c}}

\newcommand{\note}[1]{{\color{darkred}#1}}
\definecolor{darkred}{rgb}{0.65,0,0}

\newcommand{\SCICH}{\textit{SCICH}}
\newcommand{\TDCR}{\textit{TDCR}}

\newcommand{\F}{{\cal F}}

\newcommand{\Bel}{\textit{Bel}}
\newcommand{\Plaus}{\textit{Plaus}}

\newcommand{\red}{\textit{red}}
\newcommand{\blue}{\textit{blue}}
\newcommand{\yellow}{\textit{yellow}}
\newcommand{\hep}{\textit{hep}}
\newcommand{\cirr}{\textit{cirr}}
\newcommand{\gall}{\textit{gall}}
\newcommand{\pan}{\textit{pan}}

\newcommand{\sat}{\models}
\newcommand{\tendsto}{\rightarrow}
\newcommand{\Pa}{\mathit{Par}}
\newcommand{\Compat}{\mathit{Compat}}
\newcommand{\lem}{\begin{lemma}}
\newcommand{\elem}{\end{lemma}}
\newcommand{\epro}{\end{proposition}}

\newcommand{\dfn}{\begin{definition}}
\newcommand{\edfn}{\end{definition}}

\newcommand{\eprf}{\bbox\fullv{\vspace{0.1in}}}

\newcommand{\xam}{\begin{example}}
\newcommand{\exam}{\end{example}}
\renewcommand{\citeyear}{\cite}

\section{Introduction}

\commentout{
Quantitative:
Point 19: `` the likely differences in the seriousness, probability and scale of the harm or possible negative consequences.

Point 28: ``assessing the severity of the harm that an AI system can cause''

Point	 32: ``taking into account both the severity of the possible harm and its probability of occurrence''

Article 5, which outlines prohibitions and rules on the use of AI systems:

Prohibitions:
 (a) ``the placing on the market, putting into service or use of an AI system that deploys subliminal techniques beyond a person’s consciousness in order to materially distort a person’s behaviour in a manner that causes or is likely to cause that person or another person physical or psychological harm;''

Similar language for prohibition (b).

Restrictions:
``take into account ... the nature of the situation giving rise to the possible use, in particular the seriousness, probability and scale of the harm caused in the absence of the use of the system;

Article 7:
``the potential extent of such harm or such adverse impact, in particular in terms of its intensity and its ability to affect a plurality of persons;''
}


AI systems are playing an ever-expanding role in making decisions, in
applications ranging from hiring and interviewing to healthcare to
autonomous vehicles.  Perhaps not surprisingly, this is leading to
increasing scrutiny of the harm and benefit caused by (the decisions
made by) such systems.
To take just one example, the new proposal for Europe's AI
act \cite{EuropeAIAct} contains over 29 references to ``harm'' or
``harmful'', saying such things as ``\ldots it is appropriate to
classify [AI systems] as high-risk if, in the light of their intended
purpose, they pose a high risk of harm to the health and safety or the
fundamental rights of persons, taking into account both the severity
of the possible harm and its probability of
occurrence \ldots'' \cite[Proposal preamble, clause (32)]{EuropeAIAct}. 
Moreover, the European Commission recognized that if harm is to play such a
crucial role, it must be defined carefully, saying 
``Stakeholders also highlighted that \ldots it is important to define
\ldots ‘harm’ \cite[Part 2, Section 3.1]{EuropeAIAct}.
Unfortunately, defining harm appropriately has proved difficult.
Indeed, Bradley  
\citeyear{Bra12} says: 

\textit{Unfortunately, when we look at
attempts to explain the nature of harm, we find a mess.  The most
widely discussed account, the comparative account, faces
counterexamples that seem fatal.  \ldots My diagnosis is that the
notion of harm  is a Frankensteinian jumble \ldots It should be
replaced by other more well-behaved notions.}    

In \cite{BCH22a},
we defined a qualitative notion of harm (was there harm or 
wasn't there) in deterministic settings with 
no uncertainty and only a single agent, which dealt well with all
the difficulties raised in 
the philosophy literature (which also focused on qualitative harm in
deterministic settings; see \cite{CJR21} for an extensive overview).
The key features of 
our
 definition are that it 
is based on causal models and the definition of causality given by
Halpern \citeyear{Hal47,Hal48}, assumes that there is a
\emph{default} utility, and takes harm to be caused only if the
outcome has utility lower than the default.  

While getting such a definition is an important first step, it does
not address the more quantitative aspects of harm, which will clearly
be critical in comparing, for example, the harm caused by various
options, and for taking into account ``both the severity
of the possible harm and its probability of occurrence'', as suggested
in the European AI Act proposal. 
In this paper, which is an extended version of the conference paper
\cite{BCH22b}, we generalize  
our earlier definition
so as to provide a quantitative notion of harm.

The first step is relatively straightforward: we define a quantitative
notion of harm in a deterministic setting. Roughly speaking, we
take the amount of harm to be the difference between the actual
utility and the default 
utility.   Once we have this, we need
to be able to aggregate harm across different settings.  There are two
forms of aggregation that we must consider.  The first involves
dealing with uncertainty regarding the outcome.  
Here we confront issues that are well known from the decision-theory 
literature.  
There have been many rules proposed for making decisions in
the presence of uncertainty: maximizing expected utility, if
uncertainty is characterized probabilistically; \emph{maximin}
(maximizing the worst-case utility) \cite{Wald50} or \emph{minimax regret}
\cite{Niehans,Savage51} 
if there is no quantitative characterization of uncertainty;
\emph{maximin expected utility} if uncertainty is described using a set of
probability measures \cite{GS82,GS1989}.  We consider one other
approach---\emph{probability weighting}---shortly.  All of these
approaches can be applied to harm.

\commentout{
Consider the following example, discussed by Richens, Beard, and
Thompson \citeyear{RBT22} (RBT from now on): 
\xam\label{xam1} Consider two treatments for a disease which, when left untreated,
has a 50\% mortality rate. Treatment 1 has a 60\% chance of curing a
patient, and a 40\% chance of having no effect, in which case the 
disease progressing as if untreated (so that there is a 50\% mortality  
rate). Treatment 2 has an 80\% chance of curing a patient and a 20\%
chance of killing them.  As RBT point out, Treatments 1 and 2 have
identical recovery rates, yet doctors systematically favor Treatment
1, as it achieves the same recovery rate but never harms the
patient; there are no patients that would have survived had they not
been treated. On the other hand, doctors who administer Treatment 2
risk harming their patients; there are patients who die following
treatment who would have lived had they not been treated.
\exam

Using our quantitative definition of harm with the default
utility in each context to be that of doing nothing, then we get that
the expected harm of each treatment is the same.  But Treatment 1
never causes harm, while Treatment 2 has a positive probability of
causing harm.  We may want to take this into account when aggregating
harm.

More generally, when there is uncertainty about the actual context, we
confront issues that are well known from the decision-theory
literature.  There have been many rules proposed for making decisions in
the presence of uncertainty: maximizing expected utility, if
uncertainty is characterized probabilistically; \emph{maximin}
(maximizing the worst-case utility) \cite{Wald50} or \emph{minimax regret}
\cite{Niehans,Savage51} 
if there is no quantitative characterization of uncertainty;
\emph{maximin expected utility} if uncertainty is described using a set of
probability measures \cite{GS82,GS1989}.  We consider one other
approach---\emph{probability weighting}---shortly.  All of these
approaches can be applied to harm.  
}

Another issue that has received extensive
attention
in the decision-theory
literature and applies equally well to harm is that of  combining
utilities or harms of different people.  This issue arises when
we must determine the harm caused to society
of, say, a vaccination treatment, where perhaps some people will react
badly to the vaccine.  Even assuming that we can compute the harm caused
to each individual, we must consider the total harm caused to all
individuals.  An obvious approach would be to just sum up the harm
caused to each individual, but this assumes that individuals are somehow
commensurate, that is, that one person's harm of 1 should be treated
identically to another person's harm of 1.  Even if we are willing to
accept this, there is another issue to consider: fairness.  Suppose
that we have two policies, each of which cause 1 unit of harm to 1,000
people in a population of 100,000 (and cause no harm to anyone else).
We would feel quite differently about a policy if the 1,000 people to
whom harm was caused all came from a particular identifiable
population (say, poor African-Americans) than if the 1,000 people were
effectively chosen at random.

Finally, when different policies result in different probabilities of
people being harmed,
additional subtleties arise.
Heidari et al. \citeyear{HBKL21} (HBKL from now on) consider a
number of examples of 
government policies that may cause harm to each member of a
population of $n$ individuals.  (For simplicity, they assume that if
harm is caused, there is 1 unit of harm.)  Suppose that the harm
caused by a policy $P$ is characterized by the tuple $(p_1, \ldots,
p_n)$, where $p_i$ is the probability that individual $i$ suffers 1
unit of harm.  Thus, the total expected harm of policy $P$ is $p_1 + \cdots +
p_n$.  As HBKL point out and is underscored by Example~\ref{xam1},
we may feel very differently about
two policies, even if they cause the same amount of expected harm.
For example, we feel differently about a policy that necessarily harms
individual 1 and does not harm anyone else compared to a policy that
gives each individual a probability $1/n$ of being harmed.  Indeed,
there is a long line of work in psychology \cite{JenniLowenstein97}
that suggests that we find it particularly troubling to single out one victim
and concentrate all the risk of harm on him.  (This is clearly
related to the issue of unfairness to subpopulations.)  
HBKL suggest getting around these issues by aggregating harm using an
approach familiar from the decision theory literature: \emph{probability
weighting}.  The idea is to apply a weight function $w$ to the
probability and to compute the weighted expected harm.  Under the
simplifying assumption used above that, if harm is caused, it is
always 1 unit of harm, the weighted expected harm would be $w(p_1) +
\cdots  + w(p_n)$ (so we get back the standard expression for expected
harm by taking the weighting function $w$ to be the identity
(cf. \cite{Prelec98,Quiggin93}).

As HBKL point out, the policies that
are often adopted in practice seem to be the ones that optimize
weighted expected harm if we use the probability weighting functions
that empirical work has shown that people use.
HBKL take the probability function to be one that overweights small
probabilities and underweights larger probabilities.  While this works
well for their examples, the situation is actually more nuanced.  To
quote
\shortv{\cite[p. 283]{KT79}}
\fullv{Kahnemann and Tversky \citeyear[p. 283]{KT79}}
(who were the first to raise the issue):
\begin{quote}
Because people are limited in their ability to comprehend and evaluate
extreme probabilities, highly unlikely events are either neglected or
overweighted, and the difference between high probability and
certainty is either neglected or exaggerated. Thus, small
probabilities generate unpredictable behavior. Indeed, we observe two
opposite reactions to small probabilities.
\end{quote}
Indeed, as we shall see, there are examples best explained by
assuming people essentially ignore small probabilities, effectively
treating them as 0, and others that are best explained by people
overweighting small probabilities.
\commentout{
Perhaps the most common
explanation for this effect is that people underweight probabilities
when they make ``decisions from experience'' (i.e., based on their
past experience with the events of interest) and overweight
probabilities when they make ``decisions from description'' (i.e.,
the type of situation studied in a lab, where a situation is
described in words) \cite{HBWE04}. This view seems consistent with
our examples.
}

Richens, Beard, and Thompson \citeyear{RBT22} (RBT from now on) 
also proposed a quantitative and causality-based definition
of harm. 
We already discussed what we take to be problems in their approach in our
work on qualitative harm; they carry over to the quantitative
setting as well. Consider the following example that they use to
motivate their approach: 

\xam\label{xam1} Consider two treatments for a disease which, when left untreated,
has a 50\% mortality rate. Treatment 1 has a 60\% chance of curing a
patient, and a 40\% chance of having no effect, in which case the 
disease progresses as if untreated (so that there is a 50\% mortality  
rate). Treatment 2 has an 80\% chance of curing a patient and a 20\%
chance of killing them.  \fullv{As RBT point out,} Treatments 1 and 2 have
identical recovery rates, yet doctors systematically favor Treatment
1.
\exam
We agree with RBT that the explanation for this lies in the fact
that Treatment 1 causes less harm than Treatment 2.
However, we offer a different analysis that results in differences in
the degree of harm.   Specifically, for RBT,
Treatment 1 
never causes harm whereas Treatment 2 harms 10\% of all patients
(namely those patients who would have recovered had they not been
given Treatment 2). On our analysis, Treatment 1 harms 16\% of all
patients, compared to 20\% for Treatment 2. These quantitative
differences arise due to our different views on qualitative
\commentout{
harm; we leave a detailed discussion 
of this example to the \shortv{full paper \cite{BCH22b}, and
return} \fullv{appendix, and return} 
to a discussion of RBT in Section \ref{sec:com}. 
}
harm. We return to this example when discussing the differences with RBT in Section \ref{sec:com}. 
\fullv{
Briefly, we see three problems with their approach:
(1) they use but-for 
causation as their definition of causation, (2) they compare each
treatment separately to not treating the patient at all, as opposed to
 comparing 
Treatment 1, Treatment 2, and not treating the patient at all
simultaneously, and (3) they take 
as the default the outcome that would result under not treating the
patient at all.
}

The rest of the paper is organized as follows.  In
Section~\ref{sec:models} we briefly review causal models and the
definition of actual causality, since these form the basis of our
definition.
In Section~\ref{sec:harm} we provide the definition of quantitative
harm in a single context for a single agent; in
Sections~\ref{sec:harmwithuncertainty} and~\ref{sec:aggreg}, we
discuss how to extend this basic definition to situations where there
is uncertainty about the context and there are many individuals, each of which
may potentially suffer harm.  In Section~\ref{sec:benefit}, we briefly
discuss analogous definitions for benefits.  In Section~\ref{sec:com}
we compare our work to that
of RBT. As we show in Section~\ref{sec:prec}, this comparison helps to
clarify and enrich a recent lively debate regarding harm in
precision medicine. We conclude by discussing several important
avenues for future work  in Section~\ref{sec:discussion}. Finally, we
analyze the complexity of deciding and computing 
harm in the appendix. 



%
\section{Causal Models and Actual Causality
}\label{sec:models}

We start with a review of causal models 
and actual causation, 
since they play
a critical role in our definition of harm.  The material in this
section is largely taken from \cite{Hal48}.

We assume that the world is described in terms of 
variables and their values.  Some variables may have a causal
influence on others. This influence is modeled by a set of {\em
  structural equations}. It is conceptually useful to split the
variables into two sets: the {\em exogenous\/} variables, whose values
are determined by factors outside the model, and the
{\em endogenous\/} variables, whose values are ultimately determined by
the exogenous variables.  The structural equations describe how these
values are determined.

Formally, a \emph{causal model} $M$
is a pair $(\Scal,\F)$, where $\Scal$ is a \emph{signature}, which explicitly
lists the endogenous and exogenous variables  and characterizes
their possible values, and $\F$ defines a set of \emph{(modifiable)
structural equations}, relating the values of the variables.  
A signature $\Scal$ is a tuple $(\U,\V,\R)$, where $\U$ is a set of
exogenous variables, $\V$ is a set 
of endogenous variables, and $\R$ associates with every variable $Y \in 
\U \cup \V$ a nonempty set $\R(Y)$ of possible values for 
$Y$ (i.e., the set of values over which $Y$ {\em ranges}).  
For simplicity, we assume here that $\V$ is finite, as is $\R(Y)$ for
every endogenous variable $Y \in \V$.
$\F$ associates with each endogenous variable $X \in \V$ a
function denoted $F_X$
(i.e., $F_X = \F(X)$)
such that $F_X: (\times_{U \in \U} \R(U))
\times (\times_{Y \in \V - \{X\}} \R(Y)) \rightarrow \R(X)$.
This mathematical notation just makes precise the fact that 
$F_X$ determines the value of $X$,
given the values of all the other variables in $\U \cup \V$.

\commentout{
The structural equations define what happens in the presence of external
interventions. Setting the value of some set of variables $\vec{X}$ to $\vec{x}$ in a causal
model $M = (\Scal,\F)$ results in a new causal model, denoted
$M_{\vec{X}\gets \vec{x}}$, which is identical to $M$, except that the
equations for variables in $\vec{X}$ in $\F$ are replaced by $X = x$ for each $X \in \vec{X}$ and its corresponding
value $x \in \vec{x}$.
}


The dependencies between variables in a causal model $M = ((\U,\V,\R),\F)$
can be described using a {\em causal network}\index{causal
  network} (or \emph{causal graph}),
whose nodes are labeled by the endogenous and exogenous variables in
$M$, with one node for each variable in $\U \cup
\V$.  The roots of the graph are (labeled by)
the exogenous variables.  There is a directed edge from  variable $X$
to $Y$ if $Y$ \emph{depends on} $X$; this is the case
if there is some setting of all the variables in 
$\U \cup \V$ other than $X$ and $Y$ such that varying the value of
$X$ in that setting results in a variation in the value of $Y$; that
is, there is 
a setting $\vec{z}$ of the variables other than $X$ and $Y$ and values
$x$ and $x'$ of $X$ such that
$F_Y(x,\vec{z}) \ne F_Y(x',\vec{z})$.
A causal model  $M$ is \emph{recursive} (or \emph{acyclic})
if its causal graph is acyclic.
It should be clear that if $M$ is an acyclic  causal model,
then given a \emph{context}, that is, a setting $\vec{u}$ for the
exogenous variables in $\U$, the values of all the other variables are
determined (i.e., there is a unique solution to all the equations).
In this paper, following the literature, we restrict to recursive models.
We call a pair $(M,\vec{u})$ consisting of a causal model $M$ and a
context $\vec{u}$ a \emph{(causal) setting}.


A {\em causal formula (over $\Scal$)\/} is one of the form
$[Y_1 \gets y_1, \ldots, Y_k \gets y_k] \phi$,
where
$\phi$ is a Boolean combination of primitive events,
$Y_1, \ldots, Y_k$ are distinct variables in $\V$, and
$y_i \in \R(Y_i)$.
Such a formula is
abbreviated
as $[\vec{Y} \gets \vec{y}]\phi$.
The special
case where $k=0$
is abbreviated as
$\phi$.
Intuitively,
$[Y_1 \gets y_1, \ldots, Y_k \gets y_k] \phi$ says that
$\phi$ would hold if
$Y_i$ were set to $y_i$, for $i = 1,\ldots,k$.

A causal formula $\psi$ is true or false in a setting.
We write $(M,\vec{u}) \models \psi$  if
the causal formula $\psi$ is true in
the setting $(M,\vec{u})$.
The $\models$ relation is defined inductively.
$(M,\vec{u}) \models X = x$ if
the variable $X$ has value $x$
in the unique (since we are dealing with acyclic models) solution
to the equations in
$M$ in context $\vec{u}$ (that is, the
unique vector of values for the exogenous variables that simultaneously satisfies all
equations 
in $M$ 
with the variables in $\U$ set to $\vec{u}$).
Finally, 
$(M,\vec{u}) \models [\vec{Y} \gets \vec{y}]\varphi$ if 
$(M_{\vec{Y} = \vec{y}},\vec{u}) \models \varphi$,
where $M_{\vec{Y}\gets \vec{y}}$ is the causal model that is identical
to $M$, except that the 
equations for variables in $\vec{Y}$ in $\F$ are replaced by $Y = y$
for each $Y \in \vec{Y}$ and its corresponding 
value $y \in \vec{y}$.

A standard use of causal models is to define \emph{actual
causation}: that is, 
what it means for some particular event that occurred to cause 
 another particular event. 
There have been a number of definitions of actual causation given
for acyclic models
(e.g.,
\cite{beckers21c,GW07,Hall07,HP01b,Hal48,hitchcock:99,Hitchcock07,Weslake11,Woodward03}).
Although most of what we say in the remainder of the paper applies without
change to other definitions of 
actual causality in causal models, for definiteness, we focus here on
what \cite{Hal48} calls the \emph{modified}
Halpern-Pearl definition,
which we briefly review.
(See \cite{Hal48} for more intuition and motivation.) 

The events that can be causes are arbitrary conjunctions of primitive
events (formulas of the form $X=x$); 
the events that can be caused are arbitrary Boolean combinations of primitive events.  
To relate the definition of causality to the (contrastive) definition
of harm, we find it useful to give a contrastive variant of the definition
of actual causality; moreover, we are interested only in whether
$\vec{X} = \vec{x}$ causes an outcome $O=o$.  Thus,
rather than defining what it means for
$\vec{X}=\vec{x}$ to be an (actual) cause of 
an arbitrary formula $\phi$, 
we restrict ourselves to defining what it
means for $\vec{X}=\vec{x}$ \emph{rather than $\vec{X} = \vec{x}'$} to be a
cause of $O=o$ rather than $O=o'$.

\dfn\label{def:AC}
$\vec{X} = \vec{x}$ rather than $\vec{X} = \vec{x}'$ is 
an 
\emph{actual cause} of $O=o$ rather than $O=o'$ in 
$(M,\vec{u})$ if the
following three conditions hold: 
\begin{description}
\item[{\rm AC1.}]\label{ac1} $(M,\vec{u}) \models (\vec{X} = \vec{x}) \land O=o$. 
\item[{\rm AC2.}] There is a set $\vec{W}$ of variables in $\V$
and a setting $\vec{w}$ of the variables in $\vec{W}$
  such that 
$(M,\vec{u}) \models \vec{W} = \vec{w}$ and
$(M,\vec{u}) \models [\vec{X} \gets \vec{x}', \vec{W} \gets \vec{w}]O=o'$,
where $o \neq o'$.
\item[{\rm AC3.}] \label{ac3}\index{AC3}  
  \commentout{
  $\vec{X}$ is minimal; there is no strict subset $\vec{X}'$ of
    $\vec{X}$ such that $\vec{X}' = \vec{x}''$ satisfies
AC2, where $\vec{x}''$ is the restriction of
$\vec{x}$ to the variables in $\vec{X}'$.
}%
    \commentout{
     $\vec{X}$ is minimal; there is no strict subset $\vec{X}''$ of
    $\vec{X}$ such that
    there exist values for which the above conditions are satisfied.
    }
     $\vec{X}$ is minimal; there is no strict subset $\vec{X}_*$ of
    $\vec{X}$ such that 
     there exist values $\vec{x}_*$ and $\vec{x}'_*$ for $\vec{X}_*$
          satisfying AC1 and AC2.
\end{description}
\edfn
\noindent AC1 just says that $\vec{X}=\vec{x}$ cannot
be considered a cause of $O=o$ unless both $\vec{X} = \vec{x}$ and
$O=o$ actually happen.
AC3 is a minimality condition, which says
that a cause has no irrelevant conjuncts. 
To see why, note that for $\vec{X}_*=\vec{x}_*$ to satisfy AC1 requires $\vec{x}_*$ and $\vec{x}$ agreeing on all values in $\vec{X}_*$. So $\vec{X}_* = \vec{x}_*$ is the result of pruning away some conjuncts from $\vec{X} = \vec{x}$.
AC2 captures the standard
but-for condition ($\vec{X}=\vec{x}$ rather than $\vec{X} = \vec{x}'$
is a cause of $O=o$ if, had 
$\vec{X}$ been $\vec{x}'$ rather than $\vec{x}$, $O=o$
would not have happened) but allows us to apply it while keeping fixed
some variables to the value that they had in the actual 
setting $(M,\vec{u})$.
In the special case that $\vec{W} = \emptyset$, we get the standard
but-for definition of causality:  if $\vec{X} = \vec{x}$ had not
occurred (because $\vec{X}$ was $\vec{x}'$ instead)
$O=o$ would not have occurred (because it would have been $O=o'$).

\section{Quantitative Harm in a Single Context
for a Single Agent}\label{sec:harm}  

In this section, we extend the qualitative notion of harm in a given
context introduced 
in our previous work \cite{BCH22a} to a 
quantitative notion.
Both the qualitative and the quantitative notions are defined relative to a particular context in a causal utility model, which is just
like a causal model, except that there is a default utility $d$, and
it is assumed that there is a a special endogenous variable $O$ (for
outcome), whose value determines the utility.   The fact that harm is
defined relative to a given context (just like causality) means that, 
implicitly, there is no uncertainty about the context.  

Formally,  a \emph{causal utility model} is a tuple  
$M= ((\U,\V,\R),\F, {\bf u}, d)$,  where $((\U,\V,\R),\F)$ is a
causal model one of whose endogenous variables is $O$, 
${\bf u}:\R(O) \rightarrow \IR$ is a
utility function on outcomes, and $d \in \IR$ is a default utility. 
Like causation, harm is assessed relative to a setting
$(M,\vec{u})$.

\dfn\label{def:harm3}
If $\vec{X}=\vec{x}$ rather than $\vec{X}=\vec{x}'$ causes $O=o$
rather than $O=o'$ in $(M,\vec{u})$, where  $M = ((\U,\V,\R),\F, {\bf u},d)$, then the 
 {\em (quantitative) harm} to agent {\bf ag} relative to
$(\vec{X}=\vec{x}',O=o')$,
denoted $QH(M,\vec{u},\vec{X}=\vec{x}',O=o')$, 
is $\max(0,\min(d_h,{\bf
u}(o')) - {\bf u}(o)))$.  
The {\em quantitative harm} to agent {\bf ag} caused by 
  $\vec{X}$ in $(M,\vec{u})$,
denoted $QH(M,\vec{u},\vec{X})$,
is $\max_{\vec{x}',o'} QH(M,\vec{u},\vec{X}=\vec{x}',O=o')$
if there is some $\vec{x}'$ and $o'$ such that $\vec{X}=\vec{x}$ rather than
$\vec{X}=\vec{x}'$ causes $O=o$ rather than $O=o'$; if there is no
such $\vec{x}'$ and $o'$, then the
quantitative harm is taken to be 0. (Note that the values $\vec{x}$ and $o$ are uniquely determined by $(M,\vec{u})$, which is why they do not need to appear in the parametrization.)
\edfn


In other words, the quantitative harm caused by $\vec{X} = \vec{x}$
\commentout{
resulting in $O=o$ is the difference between the
utility of $O=o$ and the default utility or the maximal achievable
utility, whichever is lower 
(except that we take the harm to be 0 if this difference is negative).
}
is the maximum difference between the default utility or the utility
of the contrastive outcome, whichever is lower, and the utility of the
actual outcome  
(except that we take the harm to be 0 if this difference is negative or if $\vec{X} = \vec{x}$ did not cause the actual outcome). 
%
\commentout{
Definition \ref{def:harm3} is a generalization of a simplified version
of our definition of qualitative harm. 
(See the supplementary material for our formal definition of qualitative harm.) 
Our definition of qualitative harm
has three conditions, denoted H1--H3.  Quantitative harm as we
have defined it here is positive iff conditions H1 and H2 hold.
As is argued 
by BCH,
it is relatively rare that condition H3 plays a role,
so we have  chosen to ignore it here for ease of exposition.
}
Definition \ref{def:harm3} is a generalization of 
our
 definition of
  qualitative harm (Definition \ref{def:harm2}).
Quantitative harm as we have defined it here is
positive iff 
there is qualitative harm.%

For completeness,
we reproduce the
definition of qualitative harm, that is, whether or not there was harm,
from \cite{BCH22a}. Note that we make a distinction between
harm and {\em strict} 
harm: strict harm adds a condition -- H3 -- as a further requirement.
We argued in our companion paper that H3 rarely plays a role,
so we ignore it from here onwards for ease of exposition. However,
we could define quantitative strict harm as being identical to
quantitative harm, except that we take the quantitative strict harm to
be $0$ whenever H3 is not satisfied.

\dfn\label{def:harm2}
$\vec{X} = \vec{x}$ 
\emph{harms} {\bf ag} in $(M,\vec{u})$, where $M = ((\U,\V,\R),\F, {\bf u},d)$, 
if there exist $o\in \R(O)$ and $\vec{x}' \in \R(\vec{X})$ such that
\begin{description}
  \item[{\rm H1.}]\label{h1} ${\bf u}(O=o) < d$; and
\item[{\rm H2.}]\label{h2} there exists $o' \in \R(O)$ such that
  $\vec{X}=\vec{x}$ rather than $\vec{X}=\vec{x}'$ causes $O=o$ rather
  than $O=o'$ and  
  ${\bf u}(O=o) < {\bf u}(O=o')$.
  \end{description}
$\vec{X} = \vec{x}$  \emph{strictly harms} {\bf ag} in $(M,\vec{u})$
if, in addition,
\begin{description}
\item[{\rm H3.}]\label{h3}  
   ${\bf u}(O=o) \le {\bf u}(O=o'')$ for the unique $o'' \in \R(O)$
such that $(M,\vec{u}) \models [\vec{X} \gets \vec{x}']( O=o'')$. 
\end{description}
\edfn

\commentout{
\footnote{
In \cite{BCH22a,BCH22b} we make a distinction between harm and {\em strict}
harm:
strict harm adds a further requirement, denoted H3, to the definition
of harm.
We argued in our companion paper that H3 rarely plays a role,
so we have chosen to ignore it here for ease of exposition. However,
we could define quantitative strict harm as being identical to
quantitative harm, except that we take the quantitative strict harm to
be $0$ whenever H3 is not satisfied. (See the 
\fullv{appendix}
\shortv{full paper}
for the formal definition of both qualitative harm and strict harm, as
well as results on the complexity of computing harm.)
}}

\commentout{
The following example is a slight modification of an example that appeared in~\cite{BCH22}.
\xam
An autonomous car detects an unexpected stationary car in front of it
on a highway. It could alert the driver Bob, who would then
have to react within $10$ seconds. However, 
$10$ seconds is too long: the car will crash into the stationary car
within $8$ seconds. The autonomous car's algorithm directs it to crash
into the safety fence on the side of 
the highway, injuring Bob. Bob claims that he was harmed by the
car. Moreover, he also claims that, if alerted, he would have been able
to find a better solution that would not have resulted in his being
injured (e.g., swerving into the incoming traffic 
then back to his own lane after passing the stationary car). We assume
that if the autonomous car had done nothing and collided 
with the stationary car, both drivers would have been injured much
more severely.

Let $O$ be a three-valued variable that captures the outcome:
$O=0$ if there is no crash, so Bob is not injured at all; $O=1$ if Bob
the car crashes into the safety fence; and $O=2$ if 
the car crashing into the stationary car. 
The utility function {\bf u} is defined as ${\bf u}(O=0)=1$, ${\bf
u}(O=1)=0.8$, and ${\bf u}(O=2)=0$. 
As the system
has the built-in standard that the driver's reaction time is $10$ seconds,
from its perspective, the appropriate default utility is 
determined by the outcome that occurs when it behaves optimally and autonomously
$d_c =0.8$, hence its action, which results in utility $0.8$, does not harm
Bob (as it does not have utility less than the default).
Had the car decided to alert Bob, and had Bob not reacted on time to avoid the collision,
he would've been harmed. The harm quantity for crashing into the stationary car
is $0.8-0=0.8$.
On the other hand, from Bob's perspective, the default utility 
is $d_b = 1$: the driver expects not to be injured. Moreover, 
Bob claims that $d_b$ is achievable in the setting where he was
alerted, as he would have been able to avoid the accident. Hence,
from Bob's perspective, crashing into the safety fence harmed him in $1-0.8=0.2$.
\exam
Due to our choice of utility function, the harm
potentially caused by crashing into the stationary car is higher than
that caused by crashing into the 
fence, even if we adopt Bob's perspective.
}

\commentout{
\xam[Credit card]
Consider a credit card company that uses an AI system to decide on the credit card limit.  Alice requests a credit card with $\$10K$ limit.
Let $0$ be the utility associated with the bank denying Alice's
application, $1$ the utility associated with Alice getting the credit
card with the limit she requested; for simplicity, let $x/10,000$
denote the utility of Alice being approved for a credit card with a
limit of $x \le 10,0000$.  If we take the default utility to be 1 (by
default, people are approved for the amount they request), then if
Alice is approved with a limit of $\$1K$, the harm caused by 
the bank in $1-0.1=0.9$; similarly, if her credit card application is denied,
the harm is $1-0=1$.

On the other hand, if the bank uses a system where all new credit
cards come with the limit of $\$1K$ initially (if the application is
approved), then it seems reasonable to take the default utility to be
0.1.  In that case, 
if Alice is approved with a limit of $\$1K$, she is not harmed at all. 
\exam

\xam[Waiter]
Alice has a meal in a restaurant. The bill comes to $\$100$.
Let $O=o$ be the variable representing the tip, and let the utility be
$o/100$. That is, ${\bf u}(\$100) = 1$, and ${\bf u}(\$20) = 0.2$. 
It is customary to give $20\%$ tip, hence
it seems reasonable to take the default utility to be 0.2.  However,
Alice only has $\$5$ in her  
wallet, the restaurant accepts only cash tips, and there is no ATM
nearby.
The best achievable outcome is thus $5$, 
corresponding
to Alice giving the waiter all the cash 
she has.  According to Definition~\ref{def:quant-harm1}, if
Alice gives $\$5$, 
the waiter is not harmed. If, on the other hand, Alice gives only
$\$1$, the waiter is harmed, and the harm is $0.05-0.01=0.04$. 
\exam

\xam
An AI system (deep neural network) is used to classify MR images of suspected brain tumors. The system annotates the MRIs and passes them to the surgeon.
The possible outputs are ``no tumor'', ``operable tumor'', and ``inoperable tumor''. Bob's scan is classified as ``inoperable tumor'', and Bob is not offered a surgery.
Post-mortem, Bob's family claims that the tumor could have been operated on. Was Bob harmed? 
\exam

\xam[Homeless person]
Henry is homeless and sleeps on the street, unless he obtains $\$10$, which allows him to sleep one night in the shelter. Bob passes by Henry on the way home
from work, but decides not to give Henry $\$10$. As a result, Henry sleeps on the street, in the vicinity of a pub, and a group of
drunks leaving the pub late at night beats Henry up. Clearly Henry was harmed. We would like to say that he was harmed by the group of drunks who beat him up.
By attributing appropriate utilities to the healthy and the beaten-up states, and assuming that the default state for Henry is healthy, we indeed get that
the drunks harmed Henry. However, we would also like to say that Bob did not harm Henry. How do we do that?
\exam
}

As mentioned in the introduction, decision theory often focuses on maximizing (expected) utility. In many cases this corresponds to minimizing the 
quantitative harm, but as the following example illustrates, the two approaches can come apart even if we restrict to a
single context and a single agent. 
\commentout{
For example, it is an easy consequence of Definition \ref{def:harm3}
that if all outcomes have
utility lower than the default, and we  
are faced with two choices $X=x_1$ and $X=x_2$ that lead
to $O=o_1$ and $O=o_2$, respectively, where ${\bf u}(o_1) > {\bf
u}(o_2)$, then both maximizing utility and minimizing harm result in a
preference for $X=x_1$. Once we leave this simple setting though,
the two approaches can come apart, even if we restrict to a
single context and a single agent. The following example illustrates the differences. 
}

\xam\label{xam2}
Alice has a meal in a restaurant. The bill comes to $\$100$.
Let $O=o$ be the variable representing the tip, and let the utility be
$o/100$. That is, ${\bf u}(\$100) = 1$, and ${\bf u}(\$20) = 0.2$. 
It is customary to give a $20\%$ tip, hence
it seems reasonable to take the default utility to be 0.2.  
%
However, Alice only has $\$5$ in her  
wallet, the restaurant accepts only cash tips, and there is no ATM
nearby.
The outcome that maximizes the waiter's utility is thus for Alice to tip \$5,
corresponding
to Alice giving the waiter all the cash 
she has. 
By
Definition~\ref{def:harm3}, if
Alice gives $\$5$, 
the waiter is not harmed. If, on the other hand, Alice gives only
$\$1$, the waiter is harmed, and the harm is $0.05-0.01=0.04$, which
is the difference between the maximum utility and the actual
utility. 

Now suppose that Alice in fact has $\$30$ in her wallet. Then
Alice would maximize the waiter's utility with a tip of \$30. Yet if our goal
is to minimize harm, then any tip of \$20 or more results in a harm of 0.
\exam

While most of the examples of harm in the philosophy literature
involve only a small number of variables, when we start to consider
policy makers making large-scale decision, there could be many
variables at play. 
We thus
consider the complexity of determining
harm.  We 
fully
characterize the complexity of computing both qualitative
and quantitative harm in the appendix, and discuss the implications of
the complexity 
results.

\section{Quantitative Harm When There Is Uncertainty about Contexts}
\label{sec:harmwithuncertainty}

In general, there may be uncertainty both about the causal model
(i.e., how the world works, as described by the equations) and the
context (what is true in the world).  
\commentout{
Going back to
Example~\ref{xam1}, suppose that whether Treatment 1 cures the patient
or has no effect depends on genetic features of the patient, and
similarly for whether Treatment 2 cures or kills the patient.  
We can assume that these genetic features are part of the context.
Thus, there may be a context where Treatment 1 has no effect,
Treatment 2 kills the patient, and with no treatment the patient would
have died; and another where Treatment 1 has no
effect, Treatment 2 cures the patient, and again, with no treatment
the patient would have died.  In each context, we can
compute the harm caused by each treatment according to
Definition~\ref{def:harm2}.  We can then put a probability on
these contexts (e.g., by assuming that the effects of Treatment 1,
Treatment 2, and no treatment are independent), and compute the expected
harm of each treatment, taking the default utility in a context to be
that of doing nothing. Whatever distribution we put on contexts, it
is easy to see that in each context, Treatment 1 does no harm (i.e.,
has harm 0): either it has the same outcome as doing nothing, or it
makes things strictly better.  On the other hand, Treatment 2 causes
harm in those contexts where patients die who would have lived had
they not been treated, while causing no harm in the remaining
contexts.  Thus, any distribution on contexts that gives positive
probability to contexts where Treatment 2 results in death and doing
nothing results in the patient living results in Treatment 2 have
positive expected harm, and thus causing more harm in expectation that
Treatment 1.

In this example, considering expected harm leads to different (and
arguably more intuitive) results than considering expected utility.
The choice of default also allows us to capture important aspects of
how people evaluate harm that cannot be captured by expected utility,
as the following example shows.
}%
In decision theory, this uncertainty is usually taken into account by computing
the expected utility, where the expectation is taken with respect to a
known probability distribution. Analogously, we could define the
notion of expected quantitative harm by simply computing the product
of the quantitative harm in each causal setting and the probability of that
setting. The next example illustrates that even using this
straightforward generalization of harm already results in some
interesting differences with expected utility. 

\xam\label{ex:surg} Suppose that a doctor has a choice of either prescribing
medication ($X=1$) or performing surgery ($X=0$) on a patient.  The
medication keeps the patient stable, but does not completely cure the
patient.  Call this outcome $O=1$, and assume that it has
utility $.5$.  On the other hand, the surgery cures the patient
completely with probability $1-p$ ($O=0$, with utility 1), but has a
small probability $p$ of the patient dying ($O=2$, with utility 0), due
to factors  such as the patient’s tolerance of
anesthesia and the surgeon's skill.

The expected utility of $X=1$ is $0.5$, while the expected utility of
$X=0$ is $1-p$.  Assuming that $p < 0.5$, $X=0$ is the choice that
maximizes expected utility.  If we take the default utility to be $1$,
which is reasonable if the patient views any deviation from their
normal health as unacceptable, then the harm caused by $X=1$ is $0.5$,
while the expected harm caused by $X=0$ is $p$, so minimizing expected
harm would again lead to choosing $X=0$.  However, 
suppose that the patient has been taking the medication for some time,
and has gotten used to the treatment.  In this case, $0.5$ seems like
a reasonable choice for the default.  With this choice, $X=1$ has
expected harm 0, while $X=1$ has expected harm $0.5p$, so the choice
that minimizes expected harm is $X=1$.  
Intuitively, by taking an appropriate choice of default utility, a
harm-based approach allows us to capture the idea that one should not
risk obtaining a bad outcome when there exists an alternative that is
guaranteed to result in an outcome that is good enough.
\exam

While taking expectation is very natural, it sometimes leads to
unreasonable conclusions.

%

\xam\label{xam3} Research has shown that the probability of a fatal
accident when driving at the speed limit is $1$ in a million, and that
driving at $80\%$ of the speed limit results in $50\%$ fewer fatal
accidents than driving at the speed limit, so the probability of a
fatal accident when driving at 80\% of the speed limit is 1 in 2,000,000.
However, research has also
shown that the majority of people do drive at the speed limit, and
would prefer buying a driverless car that does so as well. Furthermore,
this preference remains even after people have been informed of these
numbers. Based on this research, the manufacturer of a driverless car
company needs to implement a policy regarding the typical driving
speed of their cars. Either cars drive at the maximum speed allowed by
the speed limit, $X=1$, or cars drive at $80\%$ of the speed limit,
$X=0$. (Obviously a more realistic model would here use a continuous
variable $X$.)  

For any given trip, there are three outcomes: $O=2$ if the driver
arrives safely at its destination in the quickest way (legally)
possible, $O=1$ if the driver arrives safely at its destination but
takes a bit more time, or $O=0$ if the car crashes and the driver
dies. For each driver, the utilities are ${\bf u}(O=2)=1$, ${\bf
u}(O=1)=0.9$, ${\bf u}(O=0)=-1,000,000$.

Maximizing expected utility results in a preference for
$X=0$, but this does not match how people react.
Taking the default utility to be 1, which seems reasonable, and
minimizing expected harm leads to the same preference.  Nor does it
help to take 
the default to be
$0.9$.
\exam

We can deal with this problem by using the idea of probability
weighting from the decision-theory literature.
We assume that agents use
a \emph{probability weighting function} $w$,
where $w: [0,1] \rightarrow [0,1]$.
In order to make use of this function, from now on we take a
causal utility model to also include a probability $\Pr$ over the
exogenous settings $\vec{u} \in \R(\U)$. (Note that, in general, there
might also be uncertainty regarding the causal equations, but for
ease of exposition, we  ignore this complication here.) 
\dfn\label{def:harm5}
The \emph{weighted expected quantitative harm (WEQH)} to agent {\bf ag}
caused by 
$\vec{X}$ relative to 
model $M$ and 
weighting function $w$ is 
$$\begin{array}{l}
WEQH(M,\vec{X},w)
= \sum_{\vec{u} \in \R(\U)}
w(\Pr(\vec{u})) QH(M,\vec{u},\vec{X}).
\end{array}$$
\edfn

\commentout{
If $\vec{X}$ reflects the implementation of a policy that we are currently deciding on, as in Example \ref{xam3}, it corresponds to an intervention on the causal model $M$. This is not the case if we are instead evaluating the harm due to an existing, possibly probabilistic, policy (or any other causal mechanism). Hence we define the quantitative harm for both cases.
}%

Applying Definition~\ref{def:harm5} to our example, taking $M$ to be a
causal model of the driving situation, we can assume for
simplicity that there are three contexts of interest: in $u_0$, the
agent does not have a fatal accident if either $X=0$ or $X=1$, in
$u_1$ he has a fatal accident if $X=1$ but not if $X=0$, and in $u_2$
he has a fatal accident if either $X=0$ or $X=1$.  We can then take
the probabilities to be $999,999 /1,000,000$ for $u_0$
and $1/2,000,000$ for both $u_1$ and $u_2$.
Deciding on a policy then amounts to determining the equation for $X$:
either we choose $X=1$, or we choose $X=0$. (Of course,  in general, more
complicated policies can be considered.) 

In practice, people tend to discount the probability of fatal
accidents; they treat it as being essentially 0.  We can capture this
by taking, for example, 
$w(1/2,000,000) = 0$ and $w(999,999/1,000,000) = 1$.
Sure enough, for this choice of
$w$, the weighted 
expected
 harm of $X=1$ is lower than that of $X=0$.

In this case, $w$ underweights the low probabilities.  But in other
cases, people overweight probabilities.  
As Gigerenzer \citeyear{Gig06} observed, after the terrorist attack on
September 11, 2001, a lot of Americans decided to reduce their air
travel and drive more, presumably because they were overweighting
the likelihood of another terrorist attack.  (As Gigerenzer points
out, the net effect was a significant increase in the number of
deaths.) 
As mentioned in the introduction, HBKL give other examples where
overweighting gives answers that seem to match how people
feel about issues.

Perhaps the most common
explanation for this effect is that people underweight probabilities
when they make ``decisions from experience'' (i.e., based on their
past experience with the events of interest), although this can flip
due to recent bad experiences (as in the case of a terrorist attack)
and overweight probabilities when they make ``decisions from
description'' (i.e., 
the type of situation studied in a lab, where a situation is
described in words) \cite{HBWE04}.   In our example, agents'
experience is that people never have fatal accidents, so they
underweight the probability.  On the other hand, if the
agent recently had a death in the family due to a fatal accident, it
is likely that he would use a $w$ that overweights these probabilities.

\commentout{
Combining probability weighting with harm also lets us deal with other
apparently paradoxical observations due to Norcross \citeyear{Norcross98}.
Norcross considers three events, where it seems that $A$ results in
more harm than $B$ 
which results in more harm than $C$ which results in more harm than
$A$, leading to an inconsistent cycle.  $A$ is the event that one person dies a premature death; $B$ is
the event of 5,000,000 people suffering moderate headaches, and $C$ is the
event that 5,000,000 people each incur a one in a million risk of
dying. 
%
Norcross claims that most people would take $A$ to involve
greater harm than $B$ and would continue to do so if we replaced
the ``5,000,000''  in $B$ by any other number. Yet clearly, if we just add up
harms,  then as long as the harm of  
a moderate headache is positive, there must be some number $N$ of
people such that $N$ people suffering moderate headaches results in greater
harm than a single premature death. Norcross further offers a scenario to argue that most people
view $C$ as involving greater harm than $A$.  Finally, Norcross provides a
scenario where $B$ seems to involve greater harm than $C$.

It is instructive to look more carefully at the precise scenarios that
Norcross considers.  To argue that $B$ involves greater harm than
$C$, Norcross considers a scenario where 5,000,000 people each drive
out to get headache medication (under the assumption that
driving results in a one in a million risk of dying).  On the other hand, to
argue that $C$  involves greater harm than $A$, Norcross considers a
scenario where 5,000,000 people in a city running some small risk of
dying from some poisonous gas.  While both scenarios involve 5,000,000 people
incurring a small risk of dying, the nature of the stories is quite
different.  In one case, the risk involves something that people do
every day---driving---which their personal experience tells them
involves  
a negligible risk of death.  On the other hand, the second scenario
involves a scary new risk (which presumably people read about, rather
than having personal experience with).  The former scenario is one
where people are likely to underweight the probability of death
(essentially treating it as 0), while the second scenario is one where
people overweight the small probability.  Thus, although the actual
probabilities are the same, the weighted probabilities are quite
different, and hence the weighted expected harm is quite different.
There is actually no cycle here.  Rather, there are two quite
different instances of $C$, for which people compute the harm very
differently.

While the use of a weighting function gives us more
descriptive accuracy, is it normatively appropriate?  Clearly, at
times, the answer is no.  When people don't buy flood insurance because
they underweight the probability of floods, this is certainly not
normatively appropriate (assuming that they can't count on the
government to bail them out). On the other hand, we would claim that
at times the weighting may even be normatively appropriate.  

For example, overweighting probabilities when hearing a description of a
poorly-understood phenomenon might simply indicate a distrust of the
probabilities or uncertainty about the true probability.  A better
representation of uncertainty might be a set of probability measures;
overweighting small probabilities might simulate the effect of
considering the worst-case expected harm (the analogue of maxmin
expected utility \cite{GS82,GS1989}).  Of course, as 
Example \ref{xam3}
points out, we might more generally consider the effect of different
decision rules; while we have assumed that the agent's uncertainty is
represented by a probability, we can certainly 
use
other representations
of uncertainty and associated decision rules, as discussed in the
introduction.  

More speculatively, the underweighting that occurs when people make
decisions from experience could itself reflect a normative
preference. Perhaps there are actions (and their consequences) with
which we have experience precisely because we consider them to be part
of our normal lives.  As a
result, we are prepared to accept higher risks resulting from such
actions than from actions (or events) which are considered abnormal or
neutral. This seems to fit well with the distinction between
scenarios $B$ and $C$ from the Norcross example: people consider the
freedom to drive to the pharmacy whenever they so choose to be part of
a normal life, whereas presumably they do not particularly value the
ability to live near a factory that produces poisonous gas. 
}
As we shall see in the next section,
combining probability weighting with harm also lets us deal with other
apparently paradoxical observations due to Norcross \citeyear{Norcross98}.

\commentout{
account for
people's preferences 
by taking $w(X=1) = 0.5 \cdot w(X=0)$, capturing the observation that
people simply discard the difference in risk of having a fatal
accident when driving at the speed limit rather than below it. The
result is that the harm based approach prefers $X=1$. (Mention that we
are not here promoting risky behavior, but merely want to capture
existing preferences and explain them by way of invoking harm.) 

Joe asked whether there's a reverse example, where we overweight probabilities and therefore harm does worse than expected utility. I'd say it's not a stretch to imagine driving through a yellow light as an example of this: it's actually not very risky, and yet people prefer a car that doesn't do it. So we can extend the example by focussing on the policy question whether or not to drive through a yellow light.
}

\section{Aggregating Harm for Different Individuals}\label{sec:aggreg}

Up to now we have considered harm for a single individual.  Further
issues arise when we try to aggregate harm across many individuals, as
we will certainly need to do when we consider societal policies.
The most  straightforward approach to determining ``societal harm''
when there are a number of individuals involved is to sum the harm
done to each individual.
By using defaults appropriately, this straightforward approach already
lets us avoid some obvious problems with maximizing expected utility.

\xam[Forced organ donation]\label{ex:organ}
Suppose that Billy is a healthy person, strolling by a
hospital. In the hospital, there are $5$ patients in need of a heart,
liver, kidney, lung, and pancreas transplant, respectively.
Suppose for simplicity that these patients will die without the
transplant, and it is not available elsewhere, while Billy will die if
these organs are harvested from him. Expected utility maximization
would suggest that saving five lives is better than saving one, so the
hospital should kidnap Billy.
On the other hand, if we take the default that Billy and each of the
patients continue in their current state of health, then harvesting
Billy's organs clearly harms Billy, while not harvesting Billy's
organs harms no one, and is thus the action that minimizes harm. 
\exam


Combining probability weighting with harm also lets us deal with other
apparently paradoxical observations due to
Norcross \citeyear{Norcross98}
that we mentioned in the previous section.
Norcross considers three events, where it seems that $A$ results in
more harm than $B$ 
which results in more harm than $C$ which results in more harm than
$A$, leading to an inconsistent cycle.  $A$ is the event that one person dies a premature death; $B$ is
the event of 5,000,000 people suffering moderate headaches, and $C$ is the
event that 5,000,000 people each incur a one in a million risk of
dying. 
%
Norcross claims that most people would take $A$ to involve
greater harm than $B$ and would continue to do so if we replaced
the ``5,000,000''  in $B$ by any other number. Yet clearly, if we just add up
harms,  then as long as the harm of  
a moderate headache is positive, there must be some number $N$ of
people such that $N$ people suffering moderate headaches results in greater
harm than a single premature death. Norcross further offers a scenario to argue that most people
view $C$ as involving greater harm than $A$.  Finally, Norcross provides a
scenario where $B$ seems to involve greater harm than $C$.

It is instructive to look more carefully at the precise scenarios that
Norcross considers.  To argue that $B$ involves greater harm than
$C$, Norcross considers a scenario where 5,000,000 people each drive
out to get headache medication (under the assumption that
driving results in a one in a million risk of dying).  On the other hand, to
argue that $C$  involves greater harm than $A$, Norcross considers a
scenario where 5,000,000 people in a city running some small risk of
dying from some poisonous gas.  While both scenarios involve 5,000,000 people
incurring a small risk of dying, the nature of the stories is quite
different.  In one case, the risk involves something that people do
every day---driving---which their personal experience tells them
involves  
a negligible risk of death.  On the other hand, the second scenario
involves a scary new risk (which presumably people read about, rather
than having personal experience with).  The former scenario is one
where people are likely to underweight the probability of death
(essentially treating it as 0), while the second scenario is one where
people overweight the small probability.  Thus, although the actual
probabilities are the same, the weighted probabilities are quite
different, and hence the weighted expected harm is quite different.
There is actually no cycle here.  Rather, there are two quite
different instances of $C$, for which people compute the harm very
differently.
\commentout{
A \emph{collective causal utility model} is a tuple $M=((\U,\V,\R),\F,{\bf u},d,{\bf AG}\G)$, where $(\U,\V,\R),\F$ is a causal model that has an endogenous outcome variable $O_{\bf ag}$ for each ${\bf ag \in AG}$, and $\G$ is a partition of {\bf AG} into groups $g \in \G$. (We here assume that all agents share the same utility function and default utility, the generalization is straightforward. Also, in general we might want to compare several possible partitions against each other, but for simplicity we here stick to a single partition.)

Define epistemic state $\K$ by simply adding $\Pr$ to $M$.

\dfn\label{def:harm6}
The \emph{unfairness} caused by $\vec{X} = \vec{x}$ relative to epistemic state $\K$
and weighting function $w$ is
$$UF(\K, \Pr,\vec{X}=\vec{x},w)=\max_{g,h} | AIH_{g} - AIH_{h} |$$.

Where $AIH_{g}= WQH_{g} / \#g$, and $WQH_{g}$ is given by
$$\begin{array}{l}
WQH_{g}=\sum_{{\bf ag} \in g} WQH_{\bf ag}(\K,\Pr,\vec{X}=\vec{x},w)
\end{array}$$

The \emph{aggregate quantitative harm (AQH)} to agents {\bf AG}
caused by $\vec{X} = \vec{x}$ relative to epistemic state $\K$,
weighting function $w$, and {\em fairness threshold} $f$, is

$$\begin{array}{l}
AQH(\K,\vec{X}=\vec{x},w) 
= \sum_g WQH_{g}
\end{array}$$

if $UF \leq f$, else:
$$\begin{array}{l}
AQH(\K,\vec{X}=\vec{x},w) 
= \# {\bf AG} \max_g AIH_{g}
\end{array}$$
\edfn
}

\commentout{
As mentioned in the introduction, HBKL give other examples where
probability weighting gives answers that seem to match how people
feel about issues.
}

But there is a whole set of other issues that arise when dealing with
societal harms: there is a concern that we may disproportionately
affect certain identifiable groups.
For example, a policy requiring certain people to work during a
pandemic may have a disproportionate impact on certain 
groups.
These groups may be the gender-based or 
ethnicity-based groups traditionally considered in the fairness
literature, but in general, they need not be.  For example, a new
freeway may 
lead to
a disproportionate harm to people living in a certain
completely integrated middle-class neighborhood.   The ``group'' might
be just an individual.  Indeed, we can see Norcross's scenario $A$ as
an instance of this phenomenon, where the group is the individual that
suffers the premature death, which is intuitively more harmful than any number of people suffering a moderate headache.
Similarly, an alternative analysis of Example \ref{ex:organ} is to view Billy as a group that would be disproportionately harmed in case his organs were harvested.

We now briefly sketch a more formal approach to computing harm that
takes this type of fairness into account.  
We assume that the groups that cannot be disproportionately harmed by
a policy must be identified in advance.  A model would
include a list of all such groups.

\dfn A \emph{collective utility model} is a tuple 
$((\U,\V,\R),\F,\Pr, A, {\bf u}_A, d_A, \G, \alpha, \beta)$, where
$((\U,\V,\R),\F)$ is a causal model, $\Pr$ is a probability on contexts,
${\bf u}_A$ and $d_A$ consist of the utility functions and default
utilities for each agent in $A$, 
$\G$ is a set of \emph{identifiable} subsets of $A$,
and $\alpha$ and $\beta$ are two additional real-valued parameters,  
to be explained shortly.  Given
 $\vec{X}$,
we can compute the (weighted expected quantitative) harm caused by $\vec{X}$ to each agent
$a \in A$, according to Definition~\ref{def:harm5}.  We then sum the
harms caused to each agent, but add a penalty $\alpha$ if some group
in $G \in \G$ is 
disproportionately harmed, where $G$ is \emph{disproportionately
harmed} if the average harm caused to the agents in $G$ is $\beta$
greater than the average harm caused to the agents in $A$.
\edfn

Intuitively,  $\G$ consists of sets of agents that should not be
disproportionately harmed. $\G$ could include, for example, all
``small'' sets of agents (say, all sets of size at most 5), since
people may consider it unfair that a small group of agents should
suffer disproportionately.
Note that we can take $\alpha$ large so that if there is a policy that
does not harm any identifiable group, it is guaranteed to be preferred
to one that does harm an identifiable group.  On the other hand, if
every policy harms some identifiable group, then we are back to
comparing policies by just summing the harm caused to individuals.


\section{Harm vs. Benefit}
\label{sec:benefit}

In many situations, we need to trade off benefits and harms.
Although many authors view benefit as the opposite 
of harm~\cite{CJR21,RBT22},
we here suggest a definition for which this is not necessarily the case, while still
allowing for the aggregation of benefits and harms.
We replace the default
value $d$ by an \emph{interval} $D = [d_h,d_b]$, where 
utility lower than $d_h$ is a harm, utility higher than $d_b$ is a
benefit, and all values within $D$ are neither harm nor benefit. 
To motivate choosing an interval rather than a single value,
we can go back to the tipping example.  We can imagine that there is
an acceptable range $[d_h,d_b]$ of tips.  Tips below $d_h$ are
unacceptable, and viewed as harms; tips above $d_b$ are 
particularly generous, and viewed as benefits.  
\commentout{
We would not expect
$d_h = d_b$, in general.
That said, when doing a cost-benefit analysis, we quite often do take
there to be a baseline, where anything below the baseline is a cost,
and anything above it is a benefit (which amounts to taking $d_h = d_b$).
}

\xam\label{xam:birthday}
Consider buying a gift for a child's birthday.
Suppose that an age-appropriate gift costing between 
$\$15$ and $\$20$ is expected if one is invited to a child's
birthday.  Thus, we can take this range to be the default.  Any gift
within this range is not considered either harmful or beneficial.
If a guest shows up without a gift, or with a bad gift (costing less
than $\$15$),
the child is harmed; on the other hand, if a guest brings an
especially expensive gift (costing more than $\$20$), the child is
benefited. 
\exam

We generalize the definition of causal utility models to include $D$ as follows. A causal utility model is a tuple  
$M= ((\U,\V,\R),\F, {\bf u},D)$, where $D = [d_h, d_b] \subseteq [0,1]$ is the default utility interval. 
Definition \ref{def:harm3} stays exactly the same, with the default
utility for harm being the lower bound of $D$. 
The qualitative and quantified definitions of benefit are symmetric to the definitions of harm, with the upper bound of $D$ taken as the default value for benefit.

In most standard cost-benefit analyses, the defaults for harm and
benefit are taken to be the same 
 (which amounts to taking $d_h = d_b$).  
 However, there is no reason that
this should be so (which suggests that perhaps the standard approach
to doing a cost-benefit analysis should be reconsidered!).  As
Example~\ref{xam:birthday} shows, it is quite natural to have $d_h < d_b$.



\commentout{
\section{Running example - versions}
Our running example examines possible policies describing the behavior of an autonomous car,
and is a modification of an example that appeared in~\cite{BCH22}.
The motivation for analyzing these policies for the potential harm is that the legislation for
harm and determining who caused harm in accidents involving autonomous vehicles is currently
underway, with the Law Commission of England and Wales and the Scottish Law
Commission recommending that drivers of self-driving cars should not be legally responsible for
crashes (they are even called ``users''); rather, the onus should 
lie with the manufacturer~\cite{LC22}. The manufacturers translate this recommendation to a standard
according to which
the driver/user does not even have to pay attention while 
at the wheel. If a complex situation arises on the road requiring
the driver's attention, the car will notify the driver, giving them
$10$ seconds to  
take control. If the driver does not react in time, the car
will flash emergency lights, slow down, and eventually stop~\cite{BBC21}.

\xam[Running example, first version]
An autonomous car detects an unexpected stationary car in front of it
on a highway. There is poor visibility, hence the stationary car is detected later than usual.
The car alerts Bob, giving him $10$ seconds to react. Unfortunately, the autonomous car crashes into
the stationary car in $5$ seconds, killing Bob.
Let $O$ be the multi-valued outcome variable for Bob, with $O=1$ standing for Bob not being injured at all, 
and $O=0$ representing Bob's death from the collision. Injuries to Bob are represented by numbers between
$0$ and $1$ according to their severity.
The utility function {\bf u} is defined as equal to the value of $O$.
Bob's family sues the autonomous car manufacturer for harming Bob, claiming that the default utility for Bob is
$1$, corresponding to Bob not being injured at all. The autonomous car manufacturer claims that under the current
policy, crashing into the stationary car is the only possible outcome. The discussion leads to changing the policy,
where the car implements the emergency protocols while alerting the driver, and does not wait for $10$ seconds.
\exam

\xam[Running example, second version]
In the same situation as before, the autonomous car flashes emergency lights and slows down, while alerting Bob.
Slowing down increases the time until the collision to $8$ seconds, and the crash is less severe, injuring Bob (but not
killing him). $O=0.7$ stands for Bob's injury.
Same question with the car manufacturer and Bob's family.
\exam

\xam[Running example, third version]
The change in policy is that the car does not alert Bob at all, if the time to accident is less than $10$ seconds.
The car could alert the driver Bob, who would then
have to react within $10$ seconds. However, 
$10$ seconds is too long: the car will crash into the stationary car
within $8$ seconds. The autonomous car's algorithm directs it to crash
into the safety fence on the side of 
the highway, injuring Bob, with $O=0.8$. Bob claims that he was harmed by the
car. Moreover, he also claims that, if alerted, he would have been able
to find a better solution that would not have resulted in his being
injured (e.g., swerving into the incoming traffic 
then back to his own lane after passing the stationary car). As illustrated in the versions above,
if the autonomous car had done nothing and collided 
with the stationary car, Bob would've been killed or injured much more severely.
\exam

\xam[Running example, fourth version]
The autonomous car detects that there is a heavy flow of cars behind it. Given the poor visibility, crashing into the stationary car
slowly would create an obstacle of two cars on the road, hence leading to more accidents. On the other hand, crashing into the
stationary car full-speed would throw both cars off the road, freeing it to the traffic.
If Bob is not killed, but moderately injured, and assuming there are sufficiently many cars behind him, this might be the solution with
the best overall utility. However, recalling that the default utility is that all drivers get to their destination not injured, we get that the
utility of other drivers in the accident is either default or lower, hence Bob's car flying off the road is a bad outcome nevertheless.
This reasoning works regardless of the number of other drivers on the road.
\exam

\xam[Harm from default actions]
It is well-known that driving at $20mp/h$ as opposed to $30mp/h$ significantly reduces the probability of killing a pedestrian in case
of a collision. Driving through the yellow light, as opposed to stopping at the intersection, increases the probability of an accident only slightly.
Yet, the harm perceived from going through the yellow light is higher than from driving at the speed limit where it is $30mp/h$.
\exam

\section{Examples}

\begin{example}[Complex systems]
If a system consists of several components, some of which are possibly human, the situation is even more complicated.
\begin{itemize} 
\item Consider a DNN that classifies images and is installed in a self-driving car. This network fails to correctly classify a pedestrian, and as
a result the car runs the pedestrian over and the pedestrian is harmed. Alternatively, if the DNN errs on the side of caution and misclassifies a pigeon as a pedestrian,
it will engage brakes, and possibly cause traffic delays.

\item We can make the example above even more complex by considering the case where the DNN recognised that there is an obstacle, but misclassified it (as, say, an animal),
and the car's AI system made a decision not to brake as not to harm the passenger inside. Who is the
cause of harm?

\item An AI system analyses brain MRI scans and locates brain tumors on the scans. The annotated scans are then passed to the
physician. Is it better to err to the side of caution (that is, to mistake benign tumors for cancerous) or the other way around? What if
the physician is able to distinguish benign tumors from cancerous by examining the scan, but sometimes misses the tumor if it is not
annotated on the image?
\end{itemize}
\end{example}

\begin{example}[Aggregated harm]
There is a mass screening program in Scotland for HPV. The screening is done by AI. Given what we know about the AI
system, is it beneficial or harmful? The factors include the percentage of teenagers with HPV, the likelihood of developing cancer if
HPV is undetected, the harm from being subjected to the test, the harm from being erroneously told that the result is positive, etc.
\end{example}

\begin{example}[Quantifying harm with different probabilities]
Examples from ``On Modeling Human Perceptions of Allocation Policies with Uncertain Outcomes'': army service, environment pollution.
\end{example}

Prediction. Consider an AI system that predicts the likelihood of transplant rejection based on a number of parameters. This prediction
is taken into account when considering whether to perform a transplantation. If the AI system makes an error, does it harm the patient? 
In one direction of error, the AI system predicts a higher likelihood than it should, and the patient does not get a transplant. In the other
direction, the AI system predicts a lower likelihood of rejection, and the patient gets a transplant, hence another patient is not getting
a transplant.

Recommender systems. These AI systems recommend the user more purchases or entertainment based on their previous choices.
This involves a classification of the previous choices and a prediction based on these choices. Harm can occur both in the classification and
in prediction stage. Misclassification is harmful (consider the recent Facebook's misclassification of videos of black men as primates).
Wrong prediction is harmful for the user (disappointing), but also can be harmful for the content creator or seller, as they are not 
recommended to the right users.
}

\commentout{
\section{Complexity}

\begin{theorem}
Given $(M,\vec{u})$, where $M = ((\U,\V,\R),\F, {\bf u},d)$, deciding whether $\vec{X} = \vec{x}$ harms {\bf ag} is D$^P_2$-complete.
Formally, the language 
\[ L_h = \{ \langle M,\vec{u}, \vec{X}, \vec{x} \rangle | \vec{X} = \vec{x} \mbox{ harms {\bf ag} in } (M,\vec{u}) \} \]
is D$^P_2$-complete.
\end{theorem}
}

\section{Comparison to RBT}\label{sec:com}
As mentioned earlier, 
RBT also proposed a quantitative and causality-based definition
of harm.  
Our previous paper outlined several objections to their
qualitative definition; here we focus instead on the 
differences in the quantitative definition.

While both we and RBT distinguish causing harm from causing a decrease in utility, and we
both have a notion of default, 
there are 
several significant differences between our approach and theirs.
Rather than having a default utility, RBT assume a default action;
in a fixed context (i.e., what we consider in Section~\ref{sec:harm}),
we could choose to take the default utility to be the utility of the default
action.  However, when computing harm, RBT only do a pairwise
comparison of the harm of a given action to the harm of the default
action, and use only but-for causation rather than the more general
definition of causation given by, say, 
\cite{Hal48}.  

This can lead to a significant
difference in the calculation of harm.
Consider Example~\ref{xam1}.
Suppose that whether Treatment 1 cures the patient 
or has no effect depends on genetic features of the patient, and
similarly for whether Treatment 2 cures or kills the patient.  
We can assume that these genetic features are part of the context.
Thus, there may be a context where Treatment 1 has no effect,
Treatment 2 kills the patient, and with no treatment the patient would
have died; and another where Treatment 1 has no
effect, Treatment 2 cures the patient, and again, with no treatment
the patient would have died. In each context, we can
compute the harm caused by each treatment. We can then put a probability on
these contexts (e.g., by assuming that the effects of Treatment 1,
Treatment 2, and no treatment are independent), and compute the expected
harm of each treatment. 

We start with the analysis by RBT, which is fairly simple. They take the default to be the outcome that results when doing nothing,
and  compare the outcome under Treatment $i$ with the outcome under the default for $i=\{1,2\}$: there is harm whenever the latter outcome is better,
so that the patient would have been cured ``but-for'' the
treatment. As there is no context in which Treatment 1 does worse than
no treatment, 
Treatment 1 never harms the patient. For Treatment 2, the relevant contexts are
those in which Treatment 2 kills the patient yet no treatment would
have cured 
the patient; these occur with probability $0.1$ (since 50\% of all
patients who are killed by Treatment 2 would have been cured without
treatment).  

There are two differences between our analysis and that of RBT that
relate purely to how 
we assess causation: instead of using but-for causation as RBT do, we use
Definition \ref{def:AC}, and instead of considering only one
counterfactual action (namely, what they take to be the \emph{default
action}), we consider all possible counterfactuals, as is 
standard. We don't need to give the details of the causal model here;
it suffices to point out that for Treatment 2 to kill a patient, it
has to preempt the effect of the disease. Therefore, in all contexts
in which a patient dies after undergoing Treatment 2, the treatment
caused the death. These contexts have probability $0.2$. 

The causal properties of Treatment 1 are different: either it cures a
patient, or it does not interfere with the disease at all. There are
two sorts of contexts in which a patient dies after undergoing
Treatment 1. In the first, the patient would have survived under
Treatment 2. Thus, Treatment 1 is a but-for cause of the patient's
death. These contexts have probability 
$0.2 \times 0.8 = 0.16$. In the
second, the patient would have been killed by Treatment 2 (and also
if they are not treated, of course); there's no event that we can hold
fixed to get a different outcome. Thus, in these contexts, Treatment 1
does not cause the patient's death, and therefore does not harm the
patient either.  

As a result, Treatment 1 has a $0.16$ probability of causing death,
compared to $0.2$ for Treatment 2. What's left is to decide on the
default utility. Given that we have such effective treatments, to us
it seems reasonable to take the default utility to be that of the patient
being cured, in which the probabilities of causing harm are simply the
probabilities of causing death.
RBT, on the other hand, take
the default 
action to be not getting treated. While this also
gets the result that Treatment 1 causes less harm than Treatment 2,
we believe that this choice of default is
highly problematic, because it has the following consequence: if we
decide not to treat any patients, we end up harming nobody! Thus, 
with this choice of default, not treating the patient 
would be preferred over Treatment 2, which 
seems inappropriate, and at odds with both our intuition and
established medical practice.

\commentout{
Moreover, assuming a default
action (in this case, that of providing no treatment) seems to lead to
an inappropriate conclusion that is avoided by using a default utility instead.
}

Going on, rather than minimizing expected harm (and possibly applying a
weighting function to the probability, as we do), when deciding which
action $a$ to perform, RBT maximize a
different utility function, namely
$U[a] - \lambda h[a]$, where $U$ is the expected utility of $a$, 
$\lambda$ is a user-dependent 
\emph{harm aversion} coefficient, and $h[a]$ is the harm as calculated
above, that is, the decrease in utility caused (in the ``but-for''
sense) by doing $a$ rather than the default action.  Not
surprisingly, this leads to quite different 
 harms than we
would calculate.

Furthermore, RBT do not pay special attention to the issues that arise
when aggregating harm, but instead 
simply compute societal harm by summing individual harms. 
As discussed in Section~\ref{sec:aggreg}, this approach
can lead to preferences that do not match those of most people.  

Finally, RBT view benefit as the opposite of harm (i.e., in the notation of
Section~\ref{sec:benefit}, they take $d_h = d_b$).  As we pointed out,
in general it seems more appropriate not to treat benefit and harm
symmetrically, and allow for a default interval.

\section{A Perspective on Harm in Precision Medicine}\label{sec:prec}

Since the publication of our conference paper and that of RBT, a
debate has erupted over causing harm in the context of precision
medicine; the major players in the debate are Dawid and
Senn \citeyear{dawid23} (DS from 
now on), Sarvet and Stensrud \citeyear{sarvet23}  (SS from
now on), and Mueller \& Pearl \citeyear{mueller23} (MP from now on).
We show that, at least as far as harm goes, the issues that arise in this
debate are in fact simplified versions of the issues we have already
discussed when comparing our approach to that of RBT.


DS and MP focus almost exclusively on a binary treatment variable $X$
and a binary outcome variable $Y$, where $x$ and $x'$, the possible
values of $X$, represent
treatment and no treatment, respectively, while $y$ and $y'$, the
possible values of $Y$, 
represent recovery and death, respectively. DS defend 
the standard practice of taking the decision rule to be ``treat the
patient if and only if the average treatment effect (ATE) is positive'',
where the ATE is defined as $ATE=Pr(y_x) - Pr(y_{x'})$.  (Here we are
using the notation of Pearl \citeyear{pearl:book2}, so that
$\Pr(\vec{y}_{\vec{x}},\ldots,\vec{z}_{\vec{w}})=\sum_{\{\vec{u} \in \R(\U)
: (M,\vec{u}) \sat
[\vec{X} \gets \vec{x}] \vec{Y}=\vec{y} \land \ldots \land
[\vec{W} \gets \vec{w}]\vec{Z}=\vec{z}\}} \Pr(\vec{u}) $.) MP, on the
other hand, argue that we should additionally take into account the
difference between the probability of benefit and that of harm, which
they define as $\Pr(B) = \Pr(y_x,y'_{x'})$ and $\Pr(H) =
\Pr(y'_x,y_{x'})$. In words: the probability of harm is the probability
that a patient would die upon receiving treatment and would recover if
not treated, and vice versa for benefit. MP interpret the medical
principle ``first do no harm''  as stating that we should give more
weight to $\Pr(H)$ than to $\Pr(B)$. Although MP do not offer an
explicit decision rule, as noted by DS they seem to suggest a rule of
the form ``treat if and only if $\Pr(B) > (1+\lambda) \Pr(H)$ for some
$\lambda \geq 0$''. Note that this means that MP are effectively interpreting
no treatment as the default action, along the lines of RBT. It easily
follows that DS's decision rule corresponds to the special case of
taking $\lambda=0$, since $\Pr(y_x) - \Pr(y_{x'}) =\Pr(B) -
\Pr(H)$. Therefore, we can view MP as suggesting a generalizaton of the
standard decision rule defended by DS that aims to incorporate the
importance of harm in addition to the importance of the most likely
outcome.  

SS concur with DS, but consider the more general setting involving
possibly non-binary treatment and outcome variables, and a utility
function over both variables. In this setting---as also pointed
out by RBT---the standard decision rule defended by DS becomes the
standard rule of maximizing the expected utility over all possible
treatments. RBT's decison rule of maximizing expected utility for an
adjusted utiltity function is in fact a generalization of MP's
decision rule to this more general setting. Indeed, as we mentioned
in Section \ref{sec:com}, MP's $\lambda$ parameter is what RBT call
the harm aversion. Furthermore, RBT's Theorem 1 generalizes the
equality $\Pr(y_x) - \Pr(y_{x'}) =\Pr(B) - \Pr(H)$ to this more general
setting. Therefore, our discussion of the issues involved in RBT's
approach (including their use of the but-for definition that we
discuss in \citep{BCH22a}) extend to MP's approach as well. Similarly,
our discussion (and that of RBT) of the limitations of maximizing
expected utility extend to the approach of SS and DS.  

Lastly, we remark that the debate between DS, SS, and MP focuses exclusively
on the harm caused to a 
single individual. But healthcare policy is typically concerned
with groups rather than individuals.  This suggests that some of the
issues that we discussed earlier regarding the harm caused to groups
should also be relevant in the context of precision medicine.
We leave a more detailed examination of these issues to future work.

\section{Discussion and Conclusion}\label{sec:discussion}

We have given a formal definition of quantitative harm, based on  
our earlier definition of qualitative harm.
While the definition of quantitative harm for a single individual in a
fixed context, where there is no uncertainty, is fairly
straightforward given our definition of qualitative harm, as we have
pointed out, there are subtleties that arise when we add probability
and when we need to take into account fairness issues.  We have
suggested an approach for dealing with fairness issues, but clearly
work needs to be done to understand the extent to which it captures
how people actually deal with these issues.  For people to be
comfortable with policies enacted by, for example, government
agencies (such as the European AI Act), the formal approach will have
to be reasonably close to their heuristics.  The situation with
probability overweighting and underweighting is even more subtle.
Research has shown that people do both overweight and underweight
low-probability events (see, e.g., \cite{HBWE04,TZ17}).
We suspect that the underweighting that occurs when people make
decisions from experience could itself reflect a normative
preference. Perhaps there are actions (and their consequences) with
which we have experience precisely because we consider them to be part
of our normal lives.  As a
result, we are prepared to accept higher risks resulting from such
actions than from actions (or events) which are considered abnormal or
neutral. This seems to fit well with the distinction between
scenarios $B$ and $C$ from the Norcross example: people consider the
flexibility of being able to drive to the pharmacy whenever they so
choose to be part of 
a normal life, whereas presumably they do not particularly value the
ability to live near a factory that produces poisonous gas. 
In any case, while we do have some understanding of when overweighting
and underweighting occurs, a policy-maker  will have to weigh
normative and descriptive considerations in 
deciding how to compute societal harm;  assuming that people always
overweight low-probability events, as HKBL do, is clearly not
appropriate (although it may well be appropriate for the applications
considered by HKBL).
Although we have focused on probabilistic representations of
uncertainty, 
another direction worth exploring is
non-probabilistic representations of uncertainty.

In addition to the weighting of probabilities, quantitative harm is
influenced by the default utility: what matters is the difference
between the default utility and the utility of the outcome (rather
than just the utility of the outcome). Although we have here
argued for this view simply by showing that it
accords well with intuition for the examples discussed, recent
empirical research shows that people do seem to take into account a
context-dependent default in precisely this manner. Indeed, Rigoli et
al. \citeyear{RRDD16}
have shown 
that people make different choices when confronted with two cases that
have identical causal structure and identical probability
distributions over the possible monetary outcomes, but where the first
(second) case is stipulated to be a low-value (resp., high-value)
context. Crucially, this behaviour is observed despite the fact that
the subjects are informed that the context {\em does not} influence
the probabilities of the outcomes. Rigoli et al.~explain their
results by assuming that the context changes the utility function
itself, but their experiments can just as easily be explained by
assuming that the context changes the default utility instead: in a
low-value context, people take the default utility to be lower than in
a high-value context, and therefore the same utility results in
different amounts of quantitative harm for each context. It would be
interesting to construct experiments that 
can
 distinguish between
the two proposals. 

\commentout{
Finally, it is worth mentioning complexity considerations. 
As with most concepts involving actual causality, deciding whether
harm occurred is intractable  even in the single-agent 
qualitative case where there is no uncertainty. 
 In fact, we prove that harm has the same complexity as causality in
 the full paper, that is, DP-complete~\cite{BCH22b}.
 Adding quantitative considerations 
 results in completeness in the matching complexity class of functional problems.
 
 That said, we do not believe that
 in practice, complexity considerations will be
 a major impediment to applying these definitions.  In many cases of
 interest,  the set of variables and their possible values is small.
 Exhaustive search is polynomial in the set of combinations of
 possible values of the variables, so the problem will be polynomial
 time in this case.  Furthermore, if we consider but-for causality
 (i.e., take $\vec{W} = \emptyset$ in AC2), which often suffices, then
 the problem becomes polynomial time in the 
 number of
 combinations of possible values of  $\vec{X}$.  
}
 
This
paper (and our previous one) constitute only a first
step towards providing a formal approach for determining harm in practice.
Clearly more work needs to be done, ranging from investigating whether
other elements need to be added to our framework; doing both empirical and
philosophical studies on the concrete factors that determine the
default utility, the weighting function, and the fairness parameters;
and investigating 
complexity issues more carefully. 

Moreover, there is obviously a close connection between harm and
blame, in that one is usually blameworthy for an outcome only if that
outcome constitutes a harm. Yet, unlike blame, harm does not always
contain a moral dimension, since natural events can also cause
harm. Therefore it is worthwhile to develop an account that integrates
both harm and blame into a full theory of moral responsibility. 
 We
  believe 
  that this paper already provides a
rich and useful framework, one that will be critical
for dealing with the ethical and regulatory issues of deploying AI systems.



\appendix

\commentout{
\section{Definition of Qualitative Harm}

\dfn\label{def:harm2}
$\vec{X} = \vec{x}$ 
\emph{harms} {\bf ag} in $(M,\vec{u})$, where $M = ((\U,\V,\R),\F, {\bf u},d)$, 
if there exist $o\in \R(O)$ and $\vec{x}' \in \R(\vec{X})$ such that
\begin{description}
  \item[{\rm H1.}]\label{h1} ${\bf u}(O=o) < d$; and
\item[{\rm H2.}]\label{h2} there exists $o' \in \R(O)$ such that
  $\vec{X}=\vec{x}$ rather than $\vec{X}=\vec{x}'$ causes $O=o$ rather
  than $O=o'$ and  
  ${\bf u}(O=o) < {\bf u}(O=o')$.
  \end{description}
$\vec{X} = \vec{x}$  \emph{strictly harms} {\bf ag} in $(M,\vec{u})$
if, in addition,
\begin{description}
\item[{\rm H3.}]\label{h3}  
   ${\bf u}(O=o) \le {\bf u}(O=o'')$ for the unique $o'' \in \R(O)$
such that $(M,\vec{u}) \models [\vec{X} \gets \vec{x}']( O=o'')$. 
\end{description}
\edfn
}

\section{The Complexity of Computing Harm}\label{sec:comp}

In this appendix, we analyze the complexity of deciding and computing
harm.   Before getting into details, we discuss the implications of
our results.  We show that determining whether $\vec{X}=\vec{x}$ harms
{\bf ag} is DP-complete in the qualitative case; determining the
extent to which $\vec{X}=\vec{x}$ harms {\bf ag} is
$FP^{\textit{NP}[\log{n}]}$-complete.  The complexity classes
\textit{DP} and $FP^{\textit{NP}[\log{n}]}$ are defined formally
below.  For the purposes of this discussion all that is necessary to
know is that they indicate that these problems are intractable, and
likely to take time exponential in the number of variables.
 
 That said, we do not believe that
 in practice, complexity considerations will be
 a major impediment to applying these definitions.  In many cases of
 interest,  the set of variables and their possible values is small.
 Exhaustive search is polynomial in the set of combinations of
 possible values of the variables, so the problem will be polynomial
 time in this case.
Even in cases where there are many variables (e.g., large-scale policy
decisions), we suspect that symmetry considerations will result in the
problem being far more tractable than the worst-case complexity suggests.
 Finally, note that if we consider but-for causality
 (i.e., take $\vec{W} = \emptyset$ in AC2), which often suffices, then
 the problem becomes polynomial time in the 
 number of combinations of possible values of  $\vec{X}$, and thus
 quite feasible.

Turning to the technical details, 
we start with 
qualitative harm, as defined in Definition~\ref{def:harm2}, and show 
that it has the same complexity as that of actual cause.

Halpern proved that 
determining whether $\vec{X} = \vec{x}$ is an actual cause 
of $\varphi$ in $(M,\vec{u})$ (see Definition~\ref{def:AC})
is DP-complete \cite{Hal47}, 
where \textit{DP} consists of those languages $L$ for which there exist a
language $L_1$ in \textit{NP} and a language $L_2$ in co-\textit{NP} such that $L = L_1
\cap L_2$ \cite{PY}.
It is well known that \textit{DP} is at least as hard as \textit{NP} and co-\textit{NP} (and
most likely strictly harder).%
\footnote{We assume that the reader is familiar with the complexity
classe \textit{NP} and \textit{co-NP}; see \cite{Sipser96} or any other
standard introduction to the theory of computation for details.}

\begin{theorem}\label{theo:qual-harm}
Given $\vec{X} = \vec{x}$ and $(M,\vec{u})$, 
deciding whether
$\vec{X} = \vec{x}$ harms {\bf ag} in $(M,\vec{u})$ 
is  \textit{DP}-complete.
The complexity result holds for both harm and strict harm.
\end{theorem}
\begin{proof}
  The proof is quite similar to the proof of \textit{DP}-completeness of actual cause.
We define the language $L$ as follows.
\[ L = \{ (M,\vec{u}, \vec{X} = \vec{x}, O=o) : \vec{X} = \vec{x} \mbox{ harms {\bf ag} in } (M,\vec{u}) \}. \]
We want to show that $L$ is \textit{DP}-complete.
As a first step, we have to find language $L_1$ and $L_2$ such that $L
= L_1 \cap L_2$, $L_1$ is in \textit{NP}, and $L_2$ is in \textit{co-NP}.

We first rewrite the definition of harm, spelling out explicitly how 
actual cause fits in.
\dfn\label{def:harmac}
$\vec{X} = \vec{x}$  \emph{harms} {\bf ag} in $(M,\vec{u})$, where $M = ((\U,\V,\R),\F, {\bf u},d)$, 
if there exist $o\in \R(O)$ and $\vec{x}' \in \R(\vec{X})$ such that all the following conditions hold.
\begin{description}
  \item[{\rm ACH1.}]\label{ach1} $(M,\vec{u}) \models (\vec{X} =
        \vec{x}) \land O=o$ and ${\bf u}(O=o) < d$. 
\item[{\rm ACH2.}]\label{ach2} There exists a set $\vec{W}$ of variables in $\V$ and a setting $\vec{w}$ of the variables in $\vec{W}$
  such that  $(M,\vec{u}) \models \vec{W} = \vec{w}$ and $(M,\vec{u}) \models [\vec{X} \gets \vec{x}', \vec{W} \gets \vec{w}]O=o'$,
where $o \neq o'$ and ${\bf u}(O=o) < {\bf u}(O=o')$.
\item[{\rm ACH3.}] \label{ac3'}  $\vec{X}$ is minimal; there is no
strict subset $\vec{X}_*$ of  $\vec{X}$ such that 
     there exist values $\vec{x}_*$ for $\vec{X}_*$ satisfying ACH1 and
     ACH2. 

  \end{description}
\edfn

Now let $L_1$ and $L_2$ be defined as follows:
\[ L_1 = \{ (M,\vec{u}, \vec{X} = \vec{x}, O=o): \vec{X} = \vec{x} \mbox{ satisfies ACH1 and ACH2 in Definition~\ref{def:harmac}} \}, \]
and
\[ L_2 = \{ (M,\vec{u}, \vec{X} = \vec{x}, O=o): \vec{X} = \vec{x} \mbox{ satisfies AC3 in Definition~\ref{def:harmac}} \}. \]
%
Roughly speaking, the language $L_1$
captures the existential requirements for harm, and the language $L_2$ captures the universal minimality requirement. It is easy to see
that $L = L_1 \cap L_2$.

To prove that $L$ is in \textit{DP}, it suffices to prove that $L_1$
is in \textit{NP} and $L_2$ is in \textit{co-NP}
Note that $L_1$ is similar to the corresponding language in the proof
of \textit{DP}-completeness of 
actual cause in~\cite{Hal47}. Indeed, the only difference is the two
polynomial checks ${\bf u}(O=o) < d$ and ${\bf u}(O=o) < {\bf
  u}(O=o')$
(we assume that the utility function can be computed in polynomial time).  
 Therefore, 
the fact that $L_1$ is in \textit{NP} follows from the fact
that the corresponding language in the proof of
\textit{DP}-completeness of actual cause is in \textit{NP}. 
The language $L_2$ is exactly the corresponding language in the proof
of \textit{DP}-completeness of actual cause 
(we refer the reader to \cite{Hal47} for details), 
which was shown to be in co-\textit{NP}.  
Thus, $L$ is in \textit{DP}.

The proof
that $L$
is \textit{DP}-hard proceeds by 
showing that we can reduce the problem
of 
deciding whether $\vec{X} = \vec{x}$ causes
$\varphi$
to $L$.
Given a pair $(M_0,\vec{u})$, where $M_0$ is a causal model without
a utility function, a subset $\vec{X}$ of 
endogenous variables and their values $\vec{x}$ in
$(M_0,\vec{u})$, and a Boolean formula 
$\varphi$ such that $(M_0,\vec{u}) \models \varphi$, we define a causal model $M_1$ with utility
as follows. $M_1$ contains all the variables and the structural
equations of $M_0$, plus 
a fresh (Boolean) variable $O$ whose value is determined by
$\varphi$.
Specifically, the structural equation for $O$ is such that the
parents of $O$ are the variables in $\varphi$, and if 
$\varphi$ is true then $O=1$ and if $\varphi$ is
false then $O=0$. 
The utility function 
${\bf u}$ assigns utility $0$ to $O=1$ and $1$ to $O=0$. We set the
default value $d$ to $1$. 

It is now easy to see that $\vec{X}=\vec{x}$ causes $\varphi$ in
$(M_0,\vec{u})$ iff 
$\vec{X}=\vec{x}$ harms {\bf ag} in $(M_1,\vec{u})$
(that is,  $(M_0,\vec{u}, \vec{X}=\vec{x}) \in L$).
Indeed, H1 is always satisfied, and H2 is satisfied iff $\vec{X}=\vec{x}$ causes $\varphi$ in  $(M_0,\vec{u})$.

It is  easy to see that H3, the additional requirement for strict
harm, holds if $\vec{X} = \vec{x}$ causes $\phi$ 
(and hence if $\vec{X} = \vec{x}$ harms {\bf ag}), so we get the same
complexity for harm and strict harm.
\end{proof}

Armed with this result, we can characterize the complexity of computing 
quantitative harm, as defined in Definition~\ref{def:harm3}. 
We first need to introduce another complexity class.
For a complexity class $A$, the class \fpa\ consists of all functions
that can be computed  
by a polynomial-time Turing machine with an oracle for a problem in
$A$, which on input $x$ asks a total of $O(\log{|x|})$ queries 
(cf.~\cite{Pap84}).
%
A function $f(x)$ is \fpa-\emph{hard} iff  every function $g$
in \fpa\  can be reduced to $f$, in the sense that there exist
polynomially computable functions 
$h_1$ and $h_2$ such that $g = h_2 \circ f \circ h_1$. 
A function $f$ is \fpa-\emph{complete} iff
it is in \fpa\ and is \fpa-hard.

We show that computing quantitative harm is \fp-complete. The proof
follows the same structure as the proof of the
complexity of computing the degree of responsibility in~\cite{CH04}.
Similar proof techniques were used by Krentel~\citeyear{Kre88} to
characterize the complexity of various optimization problems.

\begin{theorem}
Given $\vec{X} = \vec{x}$ and $(M,\vec{u})$,
computing $QH(M,\vec{u},\vec{X}=\vec{x})$ is
\fp-complete.
\end{theorem}
\begin{proof}
To show that the problem is in 
\fp,
we construct a
polynomial-time algorithm that computes 
$QH(M,\vec{u},\vec{X}=\vec{x})$ with $O(\log{n})$ queries to an \textit{NP} oracle 
$\mathcal{R}$.
%
An input 
to $\mathcal{R}$ 
is a tuple $\langle \vec{X} = \vec{x}, (M,\vec{u}), h \rangle$, and
the oracle answers ``YES'' 
iff $QH(M,\vec{u},\vec{X}=\vec{x}) \geq h$.
%
By Theorem~\ref{theo:qual-harm}, deciding whether
$\vec{X} = \vec{x}$ harms {\bf ag} in $(M,\vec{u})$ is in \textit{DP}.
To check whether the quantitative harm is at least $h$, we guess
a contrastive value $\vec{x}'$ of $\vec{X}$, compute the outcome $o'$
that results when $\vec{X} = \vec{x}'$, and check that $\min(d,
{\bf u}(o') - {\bf u}(o)) \ge h$.
Note that an \textit{NP}-oracle suffices for checking a \textit{DP}
question.  To see why, note that since a language $L$ in \textit{DP}
has the form $L_1 \cap L_2$, 
where $L_1$ is in \textit{NP} and $L_2$ is in co-\textit{NP}, 
we can check if $x \in L$ by checking whether $x \in L_1$ and $x \in
L_2$.  The latter two questions can be answered by an \textit{NP} oracle.

%
The algorithm proceeds as follows: It first calculates all 
possible values of harm that could be caused by $\vec{X} = \vec{x}$ in
$(M,\vec{u})$ 
by consider ${\bf u}(o) - {\bf u}(o'')$ for all values $o''$
of $O$ in $M$.
The number of values of $O$, and hence the possible values of harm,
is $O(n)$ (since the description of $M$ must include the range of $O$). 
These values are then ordered, and the algorithm performs a
binary search for 
$QH(M,\vec{u},\vec{X}=\vec{x})$, at 
each step choosing a possible value of harm such that half the
possible values of harm lie above it.   (Of course, if the number of
possible values is odd, the split is as close to half as possible.)
Given an answer to the oracle's query, the algorithm then focuses on
the appropriate half and repeats the operation.  Clearly the maximum
value can be found in 
$\lceil \log(n) \rceil$ steps.


We now prove \fp-hardness by a reduction from the problem MAXSAT, 
defined as the problem of computing the maximum number of 
variables in a given propositional Boolean formula $\varphi$ that can be assigned $1$ in a satisfying assignment for $\varphi$.
If $\varphi$ is unsatisfiable, we define MAXSAT$(\varphi)=-1$. 
We prove \fp-completeness of MAXSAT from first principles in
Theorem~\ref{theor:maxsat}.
For now, let us assume that this result and proceed
with the reduction. 

Given a propositional formula $\varphi$ over the set $\vec{A}$ of $n$
Boolean variables, we define the setting
$(M,\vec{u})$ as follows. $M = ((\U,\V,\R),\cF)$, where
\begin{itemize}
\item $\U$ consists of the single variable $E$; 
\item  $\V$ consists of the variables $\vec{A} \cup \{C,X,O\}$, 
  where $C$, $X$, and $O$ are fresh variables; 
\item $\R$ is defined so that 
each variable $S \in (V \setminus \{O\})$  has range $\R(S) = \{ 0,1 \}$,
and $O$ has range $\R(S) = \{ -1,0, \ldots, n+1 \}$;
\item $\cF$ is defined as follows:
\begin{itemize}
 \item $F_{C} = \varphi$.
 \item $F_O$ is $\sum_{A \in \vec{A}} A$ if $C = X = 1$ and is
   $-1$ otherwise.
    \item $F_S = 0$ for $S \in (\V \setminus \{O,C\})$.
\end{itemize} 
 \item The utility function $\bf{u}$ on $O$ is defined as $\bf{u}(o) =
      o$; the default utility $d$ is $n+1$. 
\end{itemize}
Let $\vec{u}$ be the context in which $E$ is set to $0$.  
Note that every variable in $\V \setminus \{O\}$ has value $0$ in
context $\vec{u}$.
In particular, all variables in $\vec{A}$ have the values $0$. We
denote by $\vec{a}$ the values 
of $\vec{A}$ in this setting. The value of $O$ in this setting is
$-1$. 

We claim that MAXSAT$(\varphi) = QH(M,\vec{u},\vec{A}=\vec{a}) - 1$. 

If $\varphi$ is unsatisfiable, then $O=-1$ under all assignments to
$\V \setminus \{O,C\}$, thus there is no harm: any assignment 
to $\vec{A}$ results in the same value of $O$, hence $QH(M,\vec{u},\vec{A}=\vec{a}) = 0$, and therefore
$QH(M,\vec{u},\vec{A}=\vec{a}) -1 = -1$, which is the value of MAXSAT$(\varphi)$ if $\varphi$ is unsatisfiable.

If $\varphi$ is satisfiable, let MAXSAT$(\varphi) = k$. Any assignment to $\vec{A}$ that has more than $k$ $1$'s results in
the value of $\varphi$ being $0$, and thus $C$ being $0$, and
$O$ being $-1$. Therefore, the maximal value of $O$ 
is reached when $\vec{A}$ is assigned the values that result in satisfying $\varphi$ with $k$ variables assigned $1$ (and $X$ is assigned
$1$ as well). This value is $k \leq n$ (note that it is still lower than the default). 
\[ QH(M,\vec{u},\vec{A}=\vec{a}) = k - (-1) = k+1, \]
and hence MAXSAT$(\varphi) = QH(M,\vec{u},\vec{A}=\vec{a}) - 1$ as required.

For the other direction, consider first the case where
$QH(M,\vec{u},\vec{A}=\vec{a}) = 0$, that is, there is no harm. Since
$O=-1$ in the 
original setting, it follows that every assignment to variables
results in $O$ being $-1$. Therefore, $C \wedge X = 0$ under 
all assignments to $\vec{A} \cup \{ X \}$, so $C=0$ under
all assignments to $\vec{A}$. It follows that $\varphi$ is
unsatisfiable. 
Then MAXSAT$(\varphi) = -1$ by definition, which is equal to
$QH(M,\vec{u},\vec{A}=\vec{a}) - 1$. 

Now consider the case when $QH(M,\vec{u},\vec{A}=\vec{a}) = k+1$ for
some $k \geq 0$. Then there exist assignments to $\vec{A}$ that 
result in $C=1$ (i.e., satisfy $\varphi$). Recall that in case
where the default value $d$ is unreachable,
$QH(M,\vec{u},\vec{A}=\vec{a})$ is 
the maximal value of $\bf{u}$ minus $\bf{u}(o)$. Since $\bf{u}(o) = -1$, we have that the maximal value of $\bf{u}$ is $k$ for some $k \geq 0$.
As $\bf{u}(o') = o'$, there exists an assignment to $\vec{A}$ under which $\varphi$ is satisfied and the number of $1$'s in this assignment
is $k$, and this is the maximal such assignment. Therefore, 
\[ MAXSAT(\varphi) = k = QH(M,\vec{u},\vec{A}=\vec{a}) -1. \] 
\end{proof}

It remains to prove that MAXSAT is \fp-complete.
\begin{theorem}\label{theor:maxsat}
MAXSAT is \fp-complete.
\end{theorem}
\begin{proof}
First we prove that  MAXSAT($\varphi$) is in \fp
by describing an algorithm in \fp for solving it.
The algorithm queries an oracle
$O_\Lan$ for the language $\Lan$, defined as follows:
\[
\Lan = \{ (\varphi,k) : k \ge 0,\, {MAXSAT}(varphi) \geq k \}.
\] 
It is easy to see that $\Lan \in NP$: a non-deterministic TM can guess an assignment that assigns at least
$k$ variables in $\varphi$ to $1$ and check whether this assignment satisfies $\varphi$.
Given $\varphi$, the
algorithm for computing MAXSAT($\varphi$)
performs a binary search for the output,
each time dividing the range of possible values by $2$ 
according to the answer of $O_\Lan$. The number of possible 
values of the output is the number of variables of $\varphi$ $+2$
(all integer values between $-1$ and the number of variables in $\varphi$ are possible),
so the algorithm asks $O_\Lan$ at most 
$\lceil \log{|\varphi|} \rceil + 1$ 
queries.

Now we show that MAXSAT is \fp-hard
by describing a generic reduction from a problem in \fp\ to MAXSAT.
Let $f:\{0,1\}^* \rightarrow \{0,1\}^*$ be a function in \fp. 
That is, there exist constants $c> 0$ and $d> 0$ and  a deterministic oracle
Turing machine $M_f$ that on input $w$ outputs $f(w)$ for each 
$w \in \Sigma^*$, and, for all sufficiently large inputs $w$, operates
in time at most $|w|^c$ and 
queries an oracle for SAT at most 
$d\log{|w|}$ times. 
We now describe a reduction from $f$ to MAXSAT. 
Since this is a reduction between function problems, we have to provide
two polynomially computable functions $h_1$ and $h_2$ such that
$f = h_1 \circ {\rm MAXSAT} \circ h_2$; that is, 
for every input
$w$, we have that $h_1(w)$ is a propositional formula $\varphi$ and
$h_2({\rm MAXSAT}(\varphi)) = f(w)$.

We start by describing a deterministic polynomial time Turing machine $M_r$
that on input $w$ computes $r(w)$. On input $w$ of length $n$, $M_r$ starts
by simulating $M_f$ on $w$. At some step during the simulation $M_f$
asks its first oracle query $q_1$. The query $q_1$ is a SAT query 
$\psi_1$. The machine $M_r$ cannot figure
out the answer to $q_1$, thus it writes $q_1$ down and continues with the 
simulation. Since $M_r$ does not know the answer, it has to simulate the run
of $M_f$ for both possible answers of the oracle.
The machine $M_r$ continues in this fashion, keeping track of all 
possible executions of $M_f$ on $w$ for each sequence of answers to the oracle
queries. Note that a sequence of answers to oracle
queries uniquely characterizes a specific execution of $M_f$ (of length 
$n^c$). Since there are $2^{d\log{n}} = n^d$ possible sequences of answers,
$M_r$ on $w$ has to simulate $n^d$
possible executions of $M_f$ on $w$, thus the running time of this step is bounded by $O(n^{c+d})$. 

The set of possible executions of $M_f$ can be viewed as a tree of
height $d \log {n}$ (ignoring the computation between queries to the
oracle).  
Note that $\log{n}$ is taken as $\log_2{n}$ in this proof.
(Since $\log_2(n) = \log(n)/\log(2)$, using $\log_2$ doesn't affect
the complexity class.)
There are at most $n^{d+1}$ queries on this tree.  These
queries have the form $\psi_i$, for $i = 1, \ldots, n^{d+1}$. 
We can assume without loss of generality that these formulas involve pairwise
disjoint sets of variables
(otherwise we can just rename variables so as to make this condition hold).

Each execution 
of $M_f$ on $w$ involves at most $d\log{n}$ queries from this collection.
Let $X_j$ be a variable that
describes the answer to the $j$th query in an  execution, for $1 \leq j \leq d\log{n}$. 
(Of course, which query $X_j$ is the answer to depends on the
execution.)
Each of the $n^d$ possible assignments to the variables $X_1, \ldots,
X_{d\log{n}}$ can be thought of as representing a number $a \in \{0, \ldots,
n^d-1\}$ in binary. 
Let $\zeta_a$ be the formula that characterizes the assignment to these
variables corresponding to the number $a$.  That is, 
$\zeta_a = \wedge_{j=1}^{d\log{n}} X^a_j$, where $X^a_j$ is 
$X_j$ if the $j$th bit of $a$ is $1$ in $a$ and 
$\neg{X_j}$ otherwise. 
Under the interpretation of $X_j$ as determining the answer to the query
$j$, each such assignment $a$ determines an execution of $M_f$.  

Note that the assignment corresponding to the highest number 
$a$ (in binary)
for which
all the queries corresponding to bits of $a$ that are $1$ are true is
the one that corresponds to the actual execution of $M_f$.
For suppose that this is not the case. 
That is, suppose that in the actual execution of $M_f$, the bits representing
answers to the oracle queries corresponds to some number $a' < a$.
Since $a' < a$, 
there exist bits on which $a$ and $a'$ disagree. Let $X_j$ be the most
significant bit on which $a$ and $a'$ disagree. Since $a' < a$, we have
that $X_j$ is $0$ in $a'$ and is $1$ in $a$.
By our choice of $a$, the query $\psi$ corresponding
to $X_j$ is a Boolean formula, thus a SAT-oracle that $M_f$ queries should
have answered ``yes'' to this query, contradicting the assumption that
$X_j = 0$.

The next formula provides the connection between the $X_i$s and the queries.
It says that if $\zeta_a$ is true, then all the queries corresponding to
the bits that are $1$ in $a$ must be Boolean formulae.
For each assignment $a$, let $a_1, \ldots, a_{N_a}$ be the queries
that were answered
``YES'' during the execution of  $M_f$ corresponding to $a$.  Note
that $N_a \le d \log{n}$. 
Define
\[ 
\eta = \bigwedge_{a=0}^{n^d-1} (\zeta_a \Rightarrow (\psi_{a_1} \wedge \ldots \psi_{a_{N_a}})).
 \]

  We can rewrite $\eta$ as
\[ \eta = \bigwedge_{a=0}^{n^d-1} (\zeta_a \Rightarrow 
           (\psi_{a_1} \wedge \ldots \wedge \psi_{a_{d\log{n}}})). \]

The idea now is to use MAXSAT to compute the highest-numbered
assignment $a$ such that all the queries corresponding to bits that are
$1$ in $a$ are true.  To do this,  it is useful to express the number $a$ in unary.  Let $U_1, \ldots, U_{n^d}$ be fresh variables. 
We introduce the following convention for representing numbers in
unary notation using these variables:~$U_1, \ldots, U_{n^d}$ 
represent a number $j$ in unary if $U_i = 1$ for each $i \leq j$ and
$U_i = 0$ for each $i > j$. 
Let $\xi$ express the fact that the $U_i$s represent
the same number as $X_1, \ldots, X_{d\log n}$, but in unary:
$$\xi=\bigwedge_{i=1}^{n^d} (\neg{U_i} \Rightarrow \neg{U_{i+1}}) \bigwedge
(\neg{U_1} \Rightarrow (X^0_1 \wedge \ldots \wedge X^0_{d\log n})) 
\wedge \bigwedge_{a=1}^{n^d-1} [ (U_a \wedge \neg{U_{a+1}}) \Rightarrow 
(X^a_1 \wedge \ldots \wedge X^a_{d\log n})].$$
The first conjunct ensures that once $U_i$s switch from $1$ to $0$,
they remain $0$; the second conjunct addresses the cawe
where the expressed number is $0$ (and hence all the $U_i$s are $0$);
the third conjunct defines the values of 
$X_1, \ldots, X_{d\log n}$ so that they represent the same number as $U_1, \ldots, U_{n^d}$, but in binary.

Let $\varphi = \eta \wedge \xi$.
Note that the size of $\varphi$ is polynomial in $n$.

We are interested in a truth assignment for
$\varphi$ that makes the most $U_i$'s true, since this will
tell us the execution actually followed by $M_f$.  
While MAXSAT($\varphi$) gives us 
the maximum number of variables that can be assigned true in
a witness for $\varphi$, 
this is not quite the information we need, as it would also include the variables of $\xi$. 
Assume for now that we have a machine that is capable of
computing a slightly generalized version of MAXSAT. We denote this
version SUBSET\_MAXSAT and define it as the maximum number of 
$1$s in a witness computed for a given subset of variables (and not
for the whole set as MAXSAT).  Formally, given a propositional Boolean formula $\psi$ 
and a subset $\vec{Z}$ of its variables, we define
 SUBSET\_MAXSAT($\psi, \vec{Z}$) 
 as the maximum number of variables in $\vec{Z}$ assigned $1$ in a
 satisfying assignment for $\psi$, or $-1$ if $\psi$ is unsatisfiable.

Clearly, SUBSET\_MAXSAT($\varphi, \{ U_1, \ldots, U_n \}$) gives us the assignment which
determines the execution of $M_f$.  Let 
$r(w) = \varphi$.  Let $s$ be the function that extracts
$f(w)$ from SUBSET\_MAXSAT($\varphi, \{ U_1, \ldots, U_n \}$).
(This can be done in polynomial time, since $f(w)$ can be computed in
polynomial time, given the answers to the oracle.)

It remains to show how to reduce SUBSET\_MAXSAT to MAXSAT.
Given a propositional Boolean formula $\psi$ over the set of variables $\vec{X}$ and
a set $\vec{Z} \subseteq \vec{X}$, we compute SUBSET\_MAXSAT($\psi, \vec{Z}$)
in the following way. Let $\vec{U} = \vec{X} \setminus \vec{Z}$. For each
$U_i \in \vec{U}$ we add a variable $U'_i$. Let $\vec{U}'$ be the set of all
the $U_i'$ variables. Define the formula $\psi'$ as
$\psi \wedge \bigwedge_{U_i \in \vec{U}} (U_i \Leftrightarrow \neg{U'_i})$. Clearly, in
all consistent assignments to $\vec{X} \cup \vec{U}'$, exactly half of
the variables 
in $\vec{U} \cup \vec{U}'$ are assigned $1$. Thus, the witness that
assigns MAXSAT($\psi'$) variables the value $1$ also assigns 
SUBSET\_MAXSAT($\psi, \vec{Z}$) variables from $\vec{Z}$ the value $1$.
The value SUBSET\_MAXSAT($\psi, \vec{Z}$) is computed by subtracting
$|\vec{U}|$ from MAXSAT($\psi'$). 
\end{proof}

\commentout{ 
For the hardness result, we proceed as follows. First, we define a 
problem MINSAT$(\varphi)$ as the \emph{minimal} number of 
variables that can be assigned $1$ in a satisfying assignment to $\varphi$.
We then prove $FP^{\textit{NP}[\log{n}]}$-completeness of
MINSAT$(\varphi)$ using 
techniques similar to the proof of $FP^{\Sigma_2[\log{n}]}$-completeness
of a related problem MINQSAT$_2(\exists \vec{X} \forall \vec{Y} \psi)$ in~\cite{CH04}.
Finally, we reduce MINSAT$(\varphi)$ to the degree of harm
using a construction similar to the one used in~\cite{CH04,ACHI17} for reducing
MINQSAT$_2(\exists \vec{X} \forall \vec{Y} \psi)$ to the degree of responsibility.
} 
\commentout{
The causal model in the reduction largely follows the structure
of $\varphi$, with the initial context $\vec{u}$ assigning $0$ to all variables of $\varphi$.
The value of a fresh (Boolean) variable $O$ determined by
the value of $\varphi$.
Specifically, if $\phi$ is true then $O=1/k$, where $k$ is the number of
variables assigned $1$, and if $\phi$ is false then $O=0$. We define ${\bf u}(o) = o$, and
the default value $d$ is $0$. Then $QH(M,\vec{u},\vec{X}=\vec{x}) =
1/$MINSAT$(\varphi)$. 
We defer the details to the full paper.
\commentout{
For the hardness result, we construct a reduction from the problem of
computing the \emph{degree of responsibility}~\cite{Hal48}. Roughly
speaking, 
the degree of responsibility measures the amount of causal influence
of
minimal-size $|\vec{X}| + |\vec{W}|$ such that $\vec{X} = \vec{x}$ is
a cause of $\phi$ with a ``witness'' $\vec{W} = \vec{w}$ 
(see condition AC2 in Definition~\ref{def:AC})  
and $X=x$ is a conjunct of $\vec{X} = \vec{x}$.
The reduction is similar to the one used in the proof of Theorem~\ref{theo:qual-harm}, except that the variable $O$ is multi-valued, with the set
of its values defined as $\{0, 1/2, 1/3, \ldots, 1/n \}$, where $n$ is
the size of the input model. If $\varphi$ is true, then
the size of the input model for which we are trying to determine the
degree of responsibility. If $\varphi$ is true, then
$O=1$. Otherwise, $O=1/k$, 
}
where $k=|\vec{X}| + |\vec{W}|$, for $\vec{W}$ demonstrating the
contrastive value false of $\varphi$. The utility function  
${\bf u}$ assigns utility $0$ to $O=1$ and $k$ to $O=1/k$.  
The default value $d$ is set to $n+1$.
It is easy to see that the degree of responsibility of $\vec{X}
= \vec{x}$ in $\varphi$ is $1/k$ iff
$QH(M,\vec{u},\vec{X}=\vec{x})=k$. 

We note that $FP^{\textit{DP}[\log{n}]}$-hardness of the degree of responsibility as defined in~\cite{Hal47} is not a published result. However, 
the proof is fairly straightforward, following the structure of the
proof of the previous definition in~\cite{ACHI17}. Namely, we
introduce a natural 
$FP^{\textit{DP}[\log{n}]}$-complete problem MAXSAT$(\varphi)$, defined as the maximal number of variables that can be assigned $1$
in a satisfying assignment to $\varphi$. This result implies $FP^{\textit{DP}[\log{n}]}$-completeness of the problem MINSAT$(\varphi)$,
defined as a \emph{minimal} number of variables that can be assigned $1$ in a satisfying assignment to $\varphi$.
Finally, the reduction from MINSAT$(\varphi)$ to the degree of responsibility follows the structure in~\cite{CH04}. We defer the details to the full version.
} 

\commentout{
For now I'm just adding the main analysis from my slides here, with some minimal explanation for you, rather than actual text that should appear in the final version. I'm not adding everything yet: here I'm just introducing the debate, so that you can follow the analysis using our approach below.

The most important articles: \cite{dawid23, sarvet23, mueller23}

Introducing the debate as conceptualized by Sarvet \& Stensrud.

General Problem Setting: Population of patients all suffering from the same condition

Several available treatments $A=a$, $\ldots$, $A=a'$

One (discrete) outcome variable $Y$

A utility function {\bf u}$(Y)$ 

Data on the effectiveness of treatments: experimental and/or observational, for specific subpopulations $\vec{X}=\vec{x}$, $\ldots$, $\vec{X}=\vec{x'}$

$$P(Y=y | do(A=a), \vec{X}=\vec{x})$$

Main Question: how do we decide which treatment to use?

{\bf Simple Setting}:

Binary outcome $Y$: $Y=0$ is recover, $Y=1$ is death

Binary treatment $A$: $A=1$ is treat, $A=0$ is not treat

Average Treatment Effect ATE (or ACE): 

$P(Y=0 | do(A=1)) - P(Y=0 | do(A=0))$

At least there is some agreement: {\bf do not treat} if  $ATE \leq 0$

{\bf Problem}: $A=0$ is considered to be the {\em default action} or the baseline policy, corresponding to ``do not treat''. What if that is not an option? This creates a problem for MP and BM.

Potential outcomes: 

Four types of patients:

\begin{enumerate}
\item $S=1$ for which $(Y^{A=1}=1$, $Y^{A=0}=0)$ (the harmed)
\item $S=2$ for which $(Y^{A=1}=0$, $Y^{A=0}=1)$ (the saved)
\item $S=3$ for which $(Y^{A=1}=1$, $Y^{A=0}=1)$
\item $S=4$ for which $(Y^{A=1}=0$, $Y^{A=0}=0)$
\end{enumerate}

{\bf Problem}: everyone so far has ignored $S=3$ under the assumption that nobody is harmed there. This is wrong: AC $\neq$ CD.

Relation between interventionist and counterfactual probabilities:

$ATE=\textcolor{blue}{P(Y=0 | do(A=1))} - \textcolor{red}{P(Y=0 | do(A=0))}$

$=\textcolor{blue}{P(Y^1=0,Y^0=0) + P(Y^1=0,Y^0=1)}$

$-  (\textcolor{red}{P(Y^1=1,Y^0=0) + P(Y^1=0,Y^0=0)})$

$=P(Y^1=0,Y^0=1) - P(Y^1=1,Y^0=0)$

$=P(S=2) - P(S=1)$

$=\text{ ``probability of benefit''} - \text{ ``probability of harm''}$

$=P(B) - P(H)$

\subsection{Minimizing Expected Harm}

Say we shift focus to a decision-rule. Then the obvious candidate is to focus on minimizing harm, as a generalization of maximizing expected utility. The following definition is simply the result of minimizing Definition \ref{def:harm5}, with everything written out explicitly. 

Actual Causation relation $AC$: $(a',y') \in AC(\vec{u},a)$ is to be interpreted as: $A=a$ rather than $A=a'$ is an actual cause of $Y=y$ rather than $Y=y'$ w.r.t. $(M,\vec{u})$.

\dfn\label{def:harm6}
The \textcolor{blue}{\emph{minimal expected harm}} caused by 
$A$ relative to 
\begin{itemize}
\item Structural Causal Model $M$, 
\item Actual Causation relation $AC$
\item utility function {\bf u}, 
\item  default utility $d_h$, and 
\item probability weighting function $w$ 
\end{itemize}
is 
$$
\min_{a} (\sum_{\vec{u}}
w(P(\vec{u})) \max_{(a',y') \in AC(\vec{u},a)} [\max(0,\min(d_h,\text{\bf
u}(y')) - \text{\bf u}(y)))])$$
\edfn

To apply this decision-rule to the standard setting from the debate, we proceed step-by-step in adding all the assumptions that occur in this debate. 

The first assumption is that the outcome variable is binary: $Y=1$ (death) or $Y=0$ (recovery). Letting $b$ and $g$ denote the utilities of the ``bad'' and ``good'' outcomes, we have $b= \text{\bf u}(Y=1) < \text{\bf u}(Y=0)=g$.

If $d_h \leq b$, then nobody is ever harmed.

If $d_h > b$, then 
\begin{enumerate}
\item if $y=0$ and $y'=0$ then we get $\max(0,\min(d_h,g) - g))=0$
\item if $y=0$ and $y'=1$ then we get $\max(0,\min(d_h,b) - g))=0$
\item if $y=1$ and $y'=0$ then we get $\max(0,\min(d_h,g) - b))=\min(d_h,g) - b > 0$
\item if $y=1$ and $y'=1$ then we get $\max(0,\min(d_h,b) - b))=0$
\end{enumerate}

If $b < g < d_h$ then $\min(d_h,g) - b = g-b=H_1$.

If $b < d_h < g$ then $\min(d_h,g) - b = d_h - b = H_2$.

Plugging this into Definition \ref{def:harm6}, we get that the \emph{minimal expected harm} caused by 
$A$ is
$$
\min_{a} \sum_{\{\vec{u} \in HS\}} P(\vec{u}) H$$
\vfill
where $H=\min(d_h,g) - b$ and 
\vfill
$HS=\{\vec{u} | A=a \text{ is an Actual Cause of } Y=1\}.$

Therefore the  \emph{most harmless treatment} is

$$
\argmin_{a} P(A=a \text{ is an Actual Cause of } Y=1).$$

The second assumption is that there is a binary treatment variable, where $A=1$ represents treatment and $A=0$ represents no treatment, so that the decision rule simply becomes:

\begin{centering}

treat 

iff

$P(A=0 \text{ is an AC of } Y=1) > P(A=1 \text{ is an AC of } Y=1)$

\end{centering}

The third assumption is the most inappropriate: assuming that  Actual Causation $=$ Counterfactual Dependence. We then get:

\begin{centering}

treat 

iff

$P(B) > P(H)$

\end{centering}

This is equivalent to: treat iff $ATE > 0$, since $ATE=P(B) - P(H)$. 

This is the decision rule defended by Dawid \& Senn, and in the setting under consideration, also the rule defended by Sarvet \& Stensrud. And yet they see it as entirely at odds with a counterfactual analysis of harm. 

\subsection{Issues}

\begin{enumerate}
\item AC $\neq$ CD, so $S=3$ should {\em not} be ignored.
 \item If we generalize to more outcomes, then $d_h$ becomes relevant. 
 \item If we generalize to multiple agents that are affected differently, then $d_h$ and  {\em group harm} ($\approx$ fairness)  becomes relevant.
\end{enumerate}

\subsubsection{Illustrating Issue 1: }

Pearl assumes we have ``treatment'' vs ``no treatment'', and $P(H)=P(S=1)$, $P(B)=P(S=2)$.

In the simple setting from above, we instead have that 
\begin{itemize}
 \item $P(H) = P(\text{treatment is an AC of } Y=1) = P(S=1) + P(S=3 \land \text{``treatment killed the patient'' }) \geq P(S=1)$,
 \item $P(B) = P(\text{no treatment is an AC of } Y=0) = P(S=2) + P(S=4 \land \text{``no treatment saved the patient'' }) = P(S=2)$.
\end{itemize}

This gives different decisions whenever 

$$0 < P(S=2) - \lambda P(S=1) <  \lambda P(S=3 \land \text{``treatment killed the patient'' })$$

\subsubsection{Illustrating Issue 2: Default Utility}

Here we can invoke our Example \ref{ex:surg}.

$A=1$: medication, $A=0$: surgery

$Y^1=1$: stable but no recovery, {\bf u}$(Y=1)=0.5$

$Y=0$: full recovery, {\bf u}$(Y=0)=1$. $P(Y^0 =0)=1-p$

$Y=2$: death, {\bf u}$(Y=0)=0$. $P(Y^0 =0)=p$

Maximize expected utility (for $p<0.5$): $A=0$

If $d_h=1$, then minimize expected harm: $A=0$.

If $d_h=0.5$, then minimize expected harm: $A=1$.

\subsubsection{Illustrating Issue 3: Multiple Agents}

Here we can invoke Example \ref{ex:organ}, but my analysis here defers from the one we currently offer. I think this one is better. 

Maximize utility: harvest Billy's organs.

Default utility for ``group'' Billy: being alive and healthy

Default utility for group patients: having a fatal condition

If we disallow disproportionate average harm for any group: do not harvest Billy's organs.
}

\section*{Acknowledgements}
The work of Sander Beckers 
was supported in part by ARO grant W911NF-17-1-0061.
The work of Hana Chockler was supported in part
by CHAI---the EPSRC Hub for Causality in Healthcare AI with Real Data (EP/Y028856/1).
The work of Joe Halpern was supported in part by 
NSF grant FMitF-2319186,
ARO grant W911NF-17-1-0061,  MURI grant W911NF-19-1-0217
from the ARO, a grant from Open Philanthropy, and a grant from the
Cooperative AI Foundation.

\bibstyle{chicago}

\end{document}